\documentclass{ws-ijmpc}

\usepackage{amsmath,graphicx,eucal,amssymb,subfigure,url}
\usepackage{setspace}
\usepackage{color}

\newtheorem{finding}{Finding}

\begin{document}  

\title{Evolving localizations in reaction-diffusion cellular automata}

\author{Andrew Adamatzky, Larry Bull}
\address{University of the West of England, Bristol, United Kingdom\\Email: \url{{andrew.adamatzky, larry.bull}@uwe.ac.uk}}

\author{Pierre Collet}
\address{LSIIT, Universit\'{e} Louis Pasteur, Strasbourg, France\\Email: \url{pierre.collet@lsiit.u-strasbg.fr}}

\author{Emmanuel Sapin}
\address{University of the West of England, Bristol, United Kingdom\\Email: \url{emmanuel2.sapin@uwe.ac.uk}}

\maketitle

\begin{abstract}

\noindent
We consider hexagonal cellular automata with immediate cell neighbourhood and three cell-states. Every cell calculates its next state depending on the integral representation of states  in its neighbourhood, i.e. how many neighbours are in each one state. 
We employ evolutionary algorithms to breed local transition functions that support mobile localizations (gliders), and characterize sets of the functions selected in terms of quasi-chemical systems. Analysis of the set of functions evolved allows to speculate that mobile localizations are likely to emerge in the quasi-chemical systems with limited diffusion of one reagent, a small number of molecules is required for amplification of travelling localizations, and reactions leading to stationary localizations involve relatively equal amount of quasi-chemical species. Techniques developed can be applied in cascading signals in nature-inspired spatially extended computing devices, and phenomenological studies and classification of non-linear discrete systems. 

\end{abstract}

\section{Introduction}

Localisations --- compact and long-living local disturbances of a medium's characteristics --- are becoming hot topic of interdisciplinary non-linear sciences~\cite{vakakis_2001,soukoulis_2001,chasottes_2005}. They can be found in almost any type of spatially extended non-linear systems, from 
liquid crystals to monomolecular arrays to reaction-diffusion chemical media~\cite{adamatzky_2003}. From a computer science point of view, localizations 
are ideal candidates for elementary processing units in 'free-space'\footnote{Term coined by Jonathan Mills, USA}, or collision-based~\cite{CBC} computing devices. The collision-based computation is rooted in logical universality of Conway's Game of Life~\cite{berlekamp}, Fredkin-Toffoli's conservative logic~\cite{fredkin:toffoli} and Margolus's physics of computation~\cite{margolus:1984}. 
In collision-based computing, quanta of information are represented by compact patterns traveling in an `empty' space
and performing computation by mutual collisions. The absence or presence, and type, of traveling patterns encode values of logical variables.
The trajectories of patterns approaching a collision site represent input variables. Trajectories of the patterns ejected from a collision represent the results of logical operations, i.e.  output variables. 

Amongst many possible natural systems supporting localizations, reaction-diffusion chemical media seem to be most appropriate candidates for 
experimental implementation of collision-based computing architectures. The medium does not require expensive or over-sophisticated equipment, 
can be handled and tuned with relative ease, and allows for visual identification of experimental results. For example, there is a particular 
type of reaction-diffusion chemical system, the Belousov-Zhabotinsky reaction in sub-excitable mode~\cite{sedina}, that supports the existence of localized wave-fragments (somewhat analogous to dissipative solitons~\cite{liehr2001}) which can play the role of the `billiard-balls'~\cite{margolus:1984} in a collision-based computing system~\cite{adamatzky_2004,ben_2005}.

In our previous works we have designed and studied a range of hexagonal cellular-automaton models of reaction-diffusion excitable chemical systems, particularly those with concentration dependent inhibition of the activator~\cite{adamatzky_hexgliders_2006,wuensche_adamatzky_2006,adamatzky_wuensche_2007}. We have analyzed three-state totalistic cellular automata on a two-dimensional lattice with hexagonal tiling, and discovered a set of specific rules that support a variety of mobile (gliders) and 
stationary (eaters) localizations, and generators of localizations (glider guns). In~\cite{adamatzky_wuensche_2007} we demonstrated that rich spatio-temporal dynamics of interacting localizations and generators of localizations can be used in implementing purposeful computation, including signal routing, multiple-valued logical operations and finite state machines. Despite success of preliminary studies, and some techniques 
developed~\cite{wuensche_2005} to pinpoint `best' rules supporting localizations, we remained somewhat puzzled and uncertain on whether rules manually selected are good representatives of a set of localization-supporting cell-state transition rules. We therefore applied the full power of evolutionary computation methods to evolve and select all possible rules that support mobile localizations in two-dimensional hexagonal ternary state cellular automata. Results of these studies are discussed in present paper.

The paper is structured as follows. In Sect.~\ref{automaton} we introduce and define ternary state hexagonal cellular automaton. The evolutionary algorithm used to breed glider-supporting rules is outlined in Sect.~\ref{algorithms}. The evolved set of glider-supporting rules is characterised in Sect.~\ref{matrices}. In Sect.~\ref{reactions} we present a set of quasi-chemical reactions derived from the evolved rules. We demonstrate, in Sect.~\ref{primitivisation}, that by integrating the most common glider-supporting rules in one set, we select the rules supporting only stationary localizations. Outcomes of the present studies are discussed in Sect.~\ref{discussions}.

\section{Reaction-diffusion hexagonal cellular automaton}
\label{automaton}

We study a totalistic cellular automaton, where a cell updates its state
depending on just the numbers, not positions, of different cell-states in 
its neighborhoods. We consider a ternary state automaton. One cell-state, $S$, 
is a dedicated substrate state: a cell in state $S$, whose neighbourhood is 
filled only with states $S$, does not change its state ($S$ is a an analogue of 
quiescent state in cellular automaton models). Two other states, $A$ and $B$, are
assigned to be reactants.

The cell-state transition rule can be written as follows: 
$$x^{t+1}=f(\sigma_A(x)^t, \sigma_B(x)^t, \sigma_S(x)^t) ,$$ 
where $\sigma_p(x)^t$ is the number of cell $x$'s neighbors with 
cell-state $p \in \{A, B, S\}$ at time step $t$.  As for
all classical cellular automata, all cells updates their states 
synchronously in discrete time-steps.  Our automata are based on a two-dimensional 
lattice with hexagonal tiling. The neighborhood size is seven: 
the central cell and its six closest neighbors.

To give a compact representation of the cell-state transition rule, we adopt the formalism
in~\cite{adamatzky_hexgliders_2006}, and represent the cell-state transition rule as a matrix
${\bf M}=(M_{ij})$, where $0 \leq i \leq j \leq 7$, $0 \leq i+j \leq 7$, and
$M_{ij} \in \{ A, B, S \}$.  The output state of each neighborhood is given by
the row-index $i$ (the number of neighbors in cell-state $A$) and column-index
$j$ (the number of neighbors in cell-state $B$). We do not have to count the
number of neighbors in cell-state $S$, because it is given by $7-(i+j)$.  A
cell with a neighborhood represented by indexes $i$ and $j$ will update to
cell-state $M_{ij}$ which can be read off the matrix. In terms of the
cell-state transition function this can be presented as follows:
$x^{t+1}=M_{\sigma_A(x)^t\sigma_B(x)^t}$.

How do matrix ${\bf M}$ entries correspond to phenomena in reaction-diffusion chemical
systems? The entries $M_{i0}=A$ and $M_{0j}=B$, $i>0$, symbolize the diffusion of reactants
$A$ and $B$. Entries $M_{ij}=p$, $p \in \{A, B, S\}$, can be interpreted as a quasi-chemical 
reaction $$iA + jB \longrightarrow p ,$$ where $i$ molecules of species $A$ react with $j$ molecules of 
species $B$ to produce species $p$. See detailed interpretations of some particular cases in~\cite{adamatzky_hexgliders_2006,wuensche_adamatzky_2006,adamatzky_wuensche_2007}.

\section{Breeding glider-supporting rules}
\label{algorithms}

We have employed evolutionary computation techniques developed by Sapin et al~\cite{sapin_2003,sapin_2003a,sapin_2004,sapin_2007} for 
evolving cellular automata which support mobile localizations (gliders). We used an evolutionary algorithm that incorporates aspects of natural selection or survival of the fittest. It maintains a population of structures (usually initially generated at random) that evolves according to rules of selection, recombination, mutation, and survival, referred to as genetic operators. A shared 'environment' is used to determine the fitness or performance of each individual in the population. The fittest individuals are more likely to be selected for reproduction through recombination and mutation. 

The search space is the set of cell-state transition rules (each rule represents a unique cellular automaton). An automaton of this space can be described by determining what will become of a cell in the next generation, depending on its neighbours. An individual is an automaton coded as a bit string of representing the values of a cell at the next generation for each neighbourhood state. A string is composed of two sub-strings. 
The first substring represents the neighbourhood states used by mobile localizations (gliders) and their values are determined by the evolution of the glider. The second substring is initialized at random. The search space contains $3^{36}$ possible cell-state transition rules.

A fitness function was computed as follows. Random configurations of cells are evolved by the tested automaton. After this evolution, the presence of gliders is checked by scanning the result of the configuration of the cells. The value of the fitness function is the number of gliders that appeared divided by the total number of cells. The 36 bits of each individual are initialized at random. The mutation function consists of mutating one bit among 36, while the recombination is a random point crossover. An elitist strategy in which the best individual of population is kept is used.
The value of the fitness function and the generation of the best rule are memorized. If after ten new generations the algorithm has not found a better rule the algorithm stops.

\section{Likehood of gliders}
\label{matrices}

From the set of evolved functions we calculated\footnote{The probabilistic matrices are calculated by Emmanuel Sapin, in the framework of EPSRC project EP/E005241/1} a set of matrices $F^z= (F_{ij})$, where $i$ is the number of neighbours in state $A$, and $j$ is the number of neighbours in state $B$, $z \in \{A, B, S \}$, which reflect likehoodness for any particular transition to contribute to glider dynamics. If one arbitrarily selects a
glider from the set of thousand gliders discovered, then the local transition function supporting the glider have transition $[i,j] \rightarrow z$ amongst its state of transitions with probability $F^z_{ij}$. We have also computed matrix $F^\#$ which indicate likehood of any particular 
state transition to be redundant, i.e. not necessary for supporting gliders.
Exact structure of the matrices is shown in Fig.~\ref{likehoodmatrices}. The matrices provide an adequate guiding choice when experimenting with randomly selected local  transition functions, see demonstration applet in~\cite{adamatzky_applet}. 

\begin{figure}
\centering
\begin{small}
\subfigure[]{
$
\mathbf{F}^S=
\begin{bmatrix}
1	 &.63&  .46&  .22&  .34&  .14&  .1&  .06&   \cr
.66&  .23&  .21&  .41&  .21&  .09&   .08&   &    \cr
.53&  .26&  .36&  .25&  .14&  .09&   &    &    \cr
.29&  .3&  .29&  .11&  .06&   &    &    &    \cr
.21&  .25&  .17&  .08&   &    &    &    &    \cr
.18&  .11&  .07&   &    &    &    &    &    \cr
.15&  .09&   &    &    &    &    &    &    \cr
.04&   &    &    &    &    &    &    &    \cr
\end{bmatrix}
$
}
\subfigure[]{
$
\mathbf{F}^A=
\begin{bmatrix}
0&   .09&   .14&  .13&  0&   0&   0&   0&   \cr
.14&  .43&  .25&  0&   0&   0&   0&   &    \cr
.12&  .27&  .23&  0&   0&   0&   &    &    \cr
.19&  .16&  0&   0&   0&   &    &    &    \cr
.15&  0&   0&   0&   &    &    &    &    \cr
0&   0&   0&   &    &    &    &    &    \cr
0&   0&   &    &    &    &    &    &    \cr
0&   &    &    &    &    &    &    &    \cr
\end{bmatrix}
$
}
\subfigure[]{
$
\mathbf{F}^B=
\begin{bmatrix}
0&   .24&  .16&  .13&  0&   0&   0&   0&   \cr
.16&  .25&  .22&  0&   0&   0&   0&   &    \cr
.2&  .24&  0&   0&   0&   0&   &    &    \cr
.15&  .05&   0&   0&   0&   &    &    &    \cr
.07&   0&   0&   0&   &    &    &    &    \cr
0&   0&   0&   &    &    &    &    &    \cr
0&   0&   &    &    &    &    &    &    \cr
0&   &    &    &    &    &    &    &    \cr
\end{bmatrix}
$
}
\subfigure[]{
$
\mathbf{F}^\#=
\begin{bmatrix}
0&   .04&   .24&  .52&  .66&  .86&  .9&  .94&  \cr
.04&   .09&   .32&  .59&  .79&  .91&  .92&  &    \cr
.15&  .23&  .41&  .75&  .86&  .91&  &    &    \cr
.37&  .49&  .71&  .89&  .94&  &    &    &    \cr
.57&  .75&  .83&  .92&  &    &    &    &    \cr
.82&  .89&  .93&  &    &    &    &    &    \cr
.85&  .91&  &    &    &    &    &    &    \cr
.96&  &    &    &    &    &    &    &    \cr
\end{bmatrix}
$
}
\end{small}
\caption{Glider-likehood matrices.}
\label{likehoodmatrices}
\end{figure}


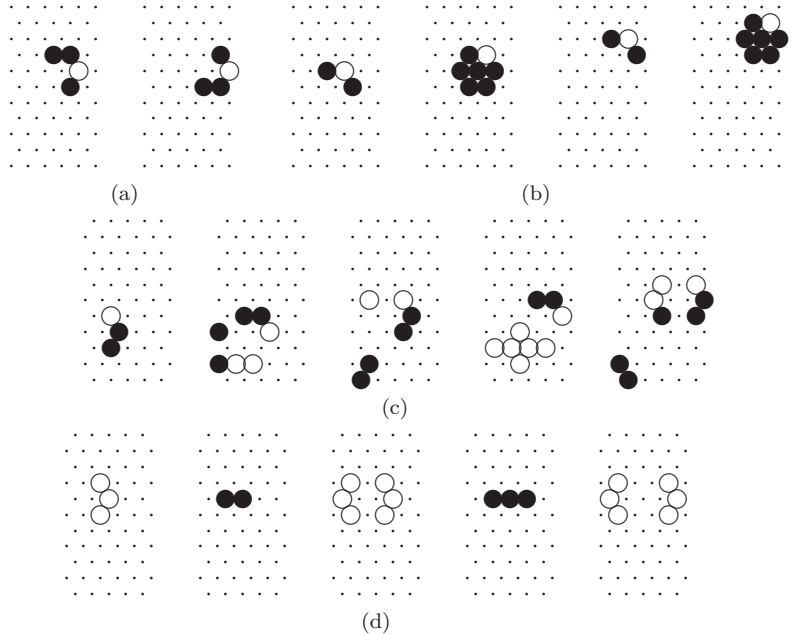
\begin{figure}
\centering
\subfigure[]{\begin{picture}(32.0,60.0)
\begin{footnotesize}
\linethickness{0.800000pt}
\put(0.0,60.0){\circle*{1.}}
\put(6.4,60.0){\circle*{1.}}
\put(12.8,60.0){\circle*{1.}}
\put(19.2,60.0){\circle*{1.}}
\put(25.6,60.0){\circle*{1.}}
\put(32.0,60.0){\circle*{1.}}
\put(3.2,54.0){\circle*{1.}}
\put(9.6,54.0){\circle*{1.}}
\put(16.0,54.0){\circle*{1.}}
\put(22.4,54.0){\circle*{1.}}
\put(28.800001,54.0){\circle*{1.}}
\put(0.0,48.0){\circle*{1.}}
\put(6.4,48.0){\circle*{1.}}
\put(12.8,48.0){\circle*{1.}}
\put(19.2,48.0){\circle*{1.}}
\put(25.6,48.0){\circle*{1.}}
\put(32.0,48.0){\circle*{1.}}
\put(3.2,42.0){\circle*{1.}}
\put(9.6,42.0){\circle*{1.}}
\put(16.0,42.0){\circle*{6.8}}
\put(22.4,42.0){\circle*{6.8}}
\put(28.800001,42.0){\circle*{1.}}
\put(0.0,36.0){\circle*{1.}}
\put(6.4,36.0){\circle*{1.}}
\put(12.8,36.0){\circle*{1.}}
\put(19.2,36.0){\circle*{1.}}
\put(25.6,36.0){\circle{6.8}}
\put(32.0,36.0){\circle*{1.}}
\put(3.2,30.0){\circle*{1.}}
\put(9.6,30.0){\circle*{1.}}
\put(16.0,30.0){\circle*{1.}}
\put(22.4,30.0){\circle*{6.8}}
\put(28.800001,30.0){\circle*{1.}}
\put(0.0,24.0){\circle*{1.}}
\put(6.4,24.0){\circle*{1.}}
\put(12.8,24.0){\circle*{1.}}
\put(19.2,24.0){\circle*{1.}}
\put(25.6,24.0){\circle*{1.}}
\put(32.0,24.0){\circle*{1.}}
\put(3.2,18.0){\circle*{1.}}
\put(9.6,18.0){\circle*{1.}}
\put(16.0,18.0){\circle*{1.}}
\put(22.4,18.0){\circle*{1.}}
\put(28.800001,18.0){\circle*{1.}}
\put(0.0,12.0){\circle*{1.}}
\put(6.4,12.0){\circle*{1.}}
\put(12.8,12.0){\circle*{1.}}
\put(19.2,12.0){\circle*{1.}}
\put(25.6,12.0){\circle*{1.}}
\put(32.0,12.0){\circle*{1.}}
\put(3.2,6.0){\circle*{1.}}
\put(9.6,6.0){\circle*{1.}}
\put(16.0,6.0){\circle*{1.}}
\put(22.4,6.0){\circle*{1.}}
\put(28.800001,6.0){\circle*{1.}}
\put(0.0,0.0){\circle*{1.}}
\put(6.4,0.0){\circle*{1.}}
\put(12.8,0.0){\circle*{1.}}
\put(19.2,0.0){\circle*{1.}}
\put(25.6,0.0){\circle*{1.}}
\put(32.0,0.0){\circle*{1.}}
\end{footnotesize}
\end{picture} \hspace{0.3cm} \begin{picture}(32.0,60.0)
\begin{footnotesize}
\linethickness{0.800000pt}
\put(0.0,60.0){\circle*{1.}}
\put(6.4,60.0){\circle*{1.}}
\put(12.8,60.0){\circle*{1.}}
\put(19.2,60.0){\circle*{1.}}
\put(25.6,60.0){\circle*{1.}}
\put(32.0,60.0){\circle*{1.}}
\put(3.2,54.0){\circle*{1.}}
\put(9.6,54.0){\circle*{1.}}
\put(16.0,54.0){\circle*{1.}}
\put(22.4,54.0){\circle*{1.}}
\put(28.800001,54.0){\circle*{1.}}
\put(0.0,48.0){\circle*{1.}}
\put(6.4,48.0){\circle*{1.}}
\put(12.8,48.0){\circle*{1.}}
\put(19.2,48.0){\circle*{1.}}
\put(25.6,48.0){\circle*{1.}}
\put(32.0,48.0){\circle*{1.}}
\put(3.2,42.0){\circle*{1.}}
\put(9.6,42.0){\circle*{1.}}
\put(16.0,42.0){\circle*{1.}}
\put(22.4,42.0){\circle*{1.}}
\put(28.800001,42.0){\circle*{6.8}}
\put(0.0,36.0){\circle*{1.}}
\put(6.4,36.0){\circle*{1.}}
\put(12.8,36.0){\circle*{1.}}
\put(19.2,36.0){\circle*{1.}}
\put(25.6,36.0){\circle*{1.}}
\put(32.0,36.0){\circle{6.8}}
\put(3.2,30.0){\circle*{1.}}
\put(9.6,30.0){\circle*{1.}}
\put(16.0,30.0){\circle*{1.}}
\put(22.4,30.0){\circle*{6.8}}
\put(28.800001,30.0){\circle*{6.8}}
\put(0.0,24.0){\circle*{1.}}
\put(6.4,24.0){\circle*{1.}}
\put(12.8,24.0){\circle*{1.}}
\put(19.2,24.0){\circle*{1.}}
\put(25.6,24.0){\circle*{1.}}
\put(32.0,24.0){\circle*{1.}}
\put(3.2,18.0){\circle*{1.}}
\put(9.6,18.0){\circle*{1.}}
\put(16.0,18.0){\circle*{1.}}
\put(22.4,18.0){\circle*{1.}}
\put(28.800001,18.0){\circle*{1.}}
\put(0.0,12.0){\circle*{1.}}
\put(6.4,12.0){\circle*{1.}}
\put(12.8,12.0){\circle*{1.}}
\put(19.2,12.0){\circle*{1.}}
\put(25.6,12.0){\circle*{1.}}
\put(32.0,12.0){\circle*{1.}}
\put(3.2,6.0){\circle*{1.}}
\put(9.6,6.0){\circle*{1.}}
\put(16.0,6.0){\circle*{1.}}
\put(22.4,6.0){\circle*{1.}}
\put(28.800001,6.0){\circle*{1.}}
\put(0.0,0.0){\circle*{1.}}
\put(6.4,0.0){\circle*{1.}}
\put(12.8,0.0){\circle*{1.}}
\put(19.2,0.0){\circle*{1.}}
\put(25.6,0.0){\circle*{1.}}
\put(32.0,0.0){\circle*{1.}}
\end{footnotesize}
\end{picture} }\hspace{0.5cm}
\subfigure[]{\begin{picture}(32.0,60.0)
\begin{footnotesize}
\linethickness{0.800000pt}
\put(0.0,60.0){\circle*{1.}}
\put(6.4,60.0){\circle*{1.}}
\put(12.8,60.0){\circle*{1.}}
\put(19.2,60.0){\circle*{1.}}
\put(25.6,60.0){\circle*{1.}}
\put(32.0,60.0){\circle*{1.}}
\put(3.2,54.0){\circle*{1.}}
\put(9.6,54.0){\circle*{1.}}
\put(16.0,54.0){\circle*{1.}}
\put(22.4,54.0){\circle*{1.}}
\put(28.800001,54.0){\circle*{1.}}
\put(0.0,48.0){\circle*{1.}}
\put(6.4,48.0){\circle*{1.}}
\put(12.8,48.0){\circle*{1.}}
\put(19.2,48.0){\circle*{1.}}
\put(25.6,48.0){\circle*{1.}}
\put(32.0,48.0){\circle*{1.}}
\put(3.2,42.0){\circle*{1.}}
\put(9.6,42.0){\circle*{1.}}
\put(16.0,42.0){\circle*{1.}}
\put(22.4,42.0){\circle*{1.}}
\put(28.800001,42.0){\circle*{1.}}
\put(0.0,36.0){\circle*{1.}}
\put(6.4,36.0){\circle*{1.}}
\put(12.8,36.0){\circle*{6.8}}
\put(19.2,36.0){\circle{6.8}}
\put(25.6,36.0){\circle*{1.}}
\put(32.0,36.0){\circle*{1.}}
\put(3.2,30.0){\circle*{1.}}
\put(9.6,30.0){\circle*{1.}}
\put(16.0,30.0){\circle*{1.}}
\put(22.4,30.0){\circle*{6.8}}
\put(28.800001,30.0){\circle*{1.}}
\put(0.0,24.0){\circle*{1.}}
\put(6.4,24.0){\circle*{1.}}
\put(12.8,24.0){\circle*{1.}}
\put(19.2,24.0){\circle*{1.}}
\put(25.6,24.0){\circle*{1.}}
\put(32.0,24.0){\circle*{1.}}
\put(3.2,18.0){\circle*{1.}}
\put(9.6,18.0){\circle*{1.}}
\put(16.0,18.0){\circle*{1.}}
\put(22.4,18.0){\circle*{1.}}
\put(28.800001,18.0){\circle*{1.}}
\put(0.0,12.0){\circle*{1.}}
\put(6.4,12.0){\circle*{1.}}
\put(12.8,12.0){\circle*{1.}}
\put(19.2,12.0){\circle*{1.}}
\put(25.6,12.0){\circle*{1.}}
\put(32.0,12.0){\circle*{1.}}
\put(3.2,6.0){\circle*{1.}}
\put(9.6,6.0){\circle*{1.}}
\put(16.0,6.0){\circle*{1.}}
\put(22.4,6.0){\circle*{1.}}
\put(28.800001,6.0){\circle*{1.}}
\put(0.0,0.0){\circle*{1.}}
\put(6.4,0.0){\circle*{1.}}
\put(12.8,0.0){\circle*{1.}}
\put(19.2,0.0){\circle*{1.}}
\put(25.6,0.0){\circle*{1.}}
\put(32.0,0.0){\circle*{1.}}
\end{footnotesize}
\end{picture} \hspace{0.3cm} \begin{picture}(32.0,60.0)
\begin{footnotesize}
\linethickness{0.800000pt}
\put(0.0,60.0){\circle*{1.}}
\put(6.4,60.0){\circle*{1.}}
\put(12.8,60.0){\circle*{1.}}
\put(19.2,60.0){\circle*{1.}}
\put(25.6,60.0){\circle*{1.}}
\put(32.0,60.0){\circle*{1.}}
\put(3.2,54.0){\circle*{1.}}
\put(9.6,54.0){\circle*{1.}}
\put(16.0,54.0){\circle*{1.}}
\put(22.4,54.0){\circle*{1.}}
\put(28.800001,54.0){\circle*{1.}}
\put(0.0,48.0){\circle*{1.}}
\put(6.4,48.0){\circle*{1.}}
\put(12.8,48.0){\circle*{1.}}
\put(19.2,48.0){\circle*{1.}}
\put(25.6,48.0){\circle*{1.}}
\put(32.0,48.0){\circle*{1.}}
\put(3.2,42.0){\circle*{1.}}
\put(9.6,42.0){\circle*{1.}}
\put(16.0,42.0){\circle*{6.8}}
\put(22.4,42.0){\circle{6.8}}
\put(28.800001,42.0){\circle*{1.}}
\put(0.0,36.0){\circle*{1.}}
\put(6.4,36.0){\circle*{1.}}
\put(12.8,36.0){\circle*{6.8}}
\put(19.2,36.0){\circle*{6.8}}
\put(25.6,36.0){\circle*{6.8}}
\put(32.0,36.0){\circle*{1.}}
\put(3.2,30.0){\circle*{1.}}
\put(9.6,30.0){\circle*{1.}}
\put(16.0,30.0){\circle*{6.8}}
\put(22.4,30.0){\circle*{6.8}}
\put(28.800001,30.0){\circle*{1.}}
\put(0.0,24.0){\circle*{1.}}
\put(6.4,24.0){\circle*{1.}}
\put(12.8,24.0){\circle*{1.}}
\put(19.2,24.0){\circle*{1.}}
\put(25.6,24.0){\circle*{1.}}
\put(32.0,24.0){\circle*{1.}}
\put(3.2,18.0){\circle*{1.}}
\put(9.6,18.0){\circle*{1.}}
\put(16.0,18.0){\circle*{1.}}
\put(22.4,18.0){\circle*{1.}}
\put(28.800001,18.0){\circle*{1.}}
\put(0.0,12.0){\circle*{1.}}
\put(6.4,12.0){\circle*{1.}}
\put(12.8,12.0){\circle*{1.}}
\put(19.2,12.0){\circle*{1.}}
\put(25.6,12.0){\circle*{1.}}
\put(32.0,12.0){\circle*{1.}}
\put(3.2,6.0){\circle*{1.}}
\put(9.6,6.0){\circle*{1.}}
\put(16.0,6.0){\circle*{1.}}
\put(22.4,6.0){\circle*{1.}}
\put(28.800001,6.0){\circle*{1.}}
\put(0.0,0.0){\circle*{1.}}
\put(6.4,0.0){\circle*{1.}}
\put(12.8,0.0){\circle*{1.}}
\put(19.2,0.0){\circle*{1.}}
\put(25.6,0.0){\circle*{1.}}
\put(32.0,0.0){\circle*{1.}}
\end{footnotesize}
\end{picture} \hspace{0.3cm} \begin{picture}(32.0,60.0)
\begin{footnotesize}
\linethickness{0.800000pt}
\put(0.0,60.0){\circle*{1.}}
\put(6.4,60.0){\circle*{1.}}
\put(12.8,60.0){\circle*{1.}}
\put(19.2,60.0){\circle*{1.}}
\put(25.6,60.0){\circle*{1.}}
\put(32.0,60.0){\circle*{1.}}
\put(3.2,54.0){\circle*{1.}}
\put(9.6,54.0){\circle*{1.}}
\put(16.0,54.0){\circle*{1.}}
\put(22.4,54.0){\circle*{1.}}
\put(28.800001,54.0){\circle*{1.}}
\put(0.0,48.0){\circle*{1.}}
\put(6.4,48.0){\circle*{1.}}
\put(12.8,48.0){\circle*{1.}}
\put(19.2,48.0){\circle*{6.8}}
\put(25.6,48.0){\circle{6.8}}
\put(32.0,48.0){\circle*{1.}}
\put(3.2,42.0){\circle*{1.}}
\put(9.6,42.0){\circle*{1.}}
\put(16.0,42.0){\circle*{1.}}
\put(22.4,42.0){\circle*{1.}}
\put(28.800001,42.0){\circle*{6.8}}
\put(0.0,36.0){\circle*{1.}}
\put(6.4,36.0){\circle*{1.}}
\put(12.8,36.0){\circle*{1.}}
\put(19.2,36.0){\circle*{1.}}
\put(25.6,36.0){\circle*{1.}}
\put(32.0,36.0){\circle*{1.}}
\put(3.2,30.0){\circle*{1.}}
\put(9.6,30.0){\circle*{1.}}
\put(16.0,30.0){\circle*{1.}}
\put(22.4,30.0){\circle*{1.}}
\put(28.800001,30.0){\circle*{1.}}
\put(0.0,24.0){\circle*{1.}}
\put(6.4,24.0){\circle*{1.}}
\put(12.8,24.0){\circle*{1.}}
\put(19.2,24.0){\circle*{1.}}
\put(25.6,24.0){\circle*{1.}}
\put(32.0,24.0){\circle*{1.}}
\put(3.2,18.0){\circle*{1.}}
\put(9.6,18.0){\circle*{1.}}
\put(16.0,18.0){\circle*{1.}}
\put(22.4,18.0){\circle*{1.}}
\put(28.800001,18.0){\circle*{1.}}
\put(0.0,12.0){\circle*{1.}}
\put(6.4,12.0){\circle*{1.}}
\put(12.8,12.0){\circle*{1.}}
\put(19.2,12.0){\circle*{1.}}
\put(25.6,12.0){\circle*{1.}}
\put(32.0,12.0){\circle*{1.}}
\put(3.2,6.0){\circle*{1.}}
\put(9.6,6.0){\circle*{1.}}
\put(16.0,6.0){\circle*{1.}}
\put(22.4,6.0){\circle*{1.}}
\put(28.800001,6.0){\circle*{1.}}
\put(0.0,0.0){\circle*{1.}}
\put(6.4,0.0){\circle*{1.}}
\put(12.8,0.0){\circle*{1.}}
\put(19.2,0.0){\circle*{1.}}
\put(25.6,0.0){\circle*{1.}}
\put(32.0,0.0){\circle*{1.}}
\end{footnotesize}
\end{picture} \hspace{0.3cm} \begin{picture}(32.0,60.0)
\begin{footnotesize}
\linethickness{0.800000pt}
\put(0.0,60.0){\circle*{1.}}
\put(6.4,60.0){\circle*{1.}}
\put(12.8,60.0){\circle*{1.}}
\put(19.2,60.0){\circle*{1.}}
\put(25.6,60.0){\circle*{1.}}
\put(32.0,60.0){\circle*{1.}}
\put(3.2,54.0){\circle*{1.}}
\put(9.6,54.0){\circle*{1.}}
\put(16.0,54.0){\circle*{1.}}
\put(22.4,54.0){\circle*{6.8}}
\put(28.800001,54.0){\circle{6.8}}
\put(0.0,48.0){\circle*{1.}}
\put(6.4,48.0){\circle*{1.}}
\put(12.8,48.0){\circle*{1.}}
\put(19.2,48.0){\circle*{6.8}}
\put(25.6,48.0){\circle*{6.8}}
\put(32.0,48.0){\circle*{6.8}}
\put(3.2,42.0){\circle*{1.}}
\put(9.6,42.0){\circle*{1.}}
\put(16.0,42.0){\circle*{1.}}
\put(22.4,42.0){\circle*{6.8}}
\put(28.800001,42.0){\circle*{6.8}}
\put(0.0,36.0){\circle*{1.}}
\put(6.4,36.0){\circle*{1.}}
\put(12.8,36.0){\circle*{1.}}
\put(19.2,36.0){\circle*{1.}}
\put(25.6,36.0){\circle*{1.}}
\put(32.0,36.0){\circle*{1.}}
\put(3.2,30.0){\circle*{1.}}
\put(9.6,30.0){\circle*{1.}}
\put(16.0,30.0){\circle*{1.}}
\put(22.4,30.0){\circle*{1.}}
\put(28.800001,30.0){\circle*{1.}}
\put(0.0,24.0){\circle*{1.}}
\put(6.4,24.0){\circle*{1.}}
\put(12.8,24.0){\circle*{1.}}
\put(19.2,24.0){\circle*{1.}}
\put(25.6,24.0){\circle*{1.}}
\put(32.0,24.0){\circle*{1.}}
\put(3.2,18.0){\circle*{1.}}
\put(9.6,18.0){\circle*{1.}}
\put(16.0,18.0){\circle*{1.}}
\put(22.4,18.0){\circle*{1.}}
\put(28.800001,18.0){\circle*{1.}}
\put(0.0,12.0){\circle*{1.}}
\put(6.4,12.0){\circle*{1.}}
\put(12.8,12.0){\circle*{1.}}
\put(19.2,12.0){\circle*{1.}}
\put(25.6,12.0){\circle*{1.}}
\put(32.0,12.0){\circle*{1.}}
\put(3.2,6.0){\circle*{1.}}
\put(9.6,6.0){\circle*{1.}}
\put(16.0,6.0){\circle*{1.}}
\put(22.4,6.0){\circle*{1.}}
\put(28.800001,6.0){\circle*{1.}}
\put(0.0,0.0){\circle*{1.}}
\put(6.4,0.0){\circle*{1.}}
\put(12.8,0.0){\circle*{1.}}
\put(19.2,0.0){\circle*{1.}}
\put(25.6,0.0){\circle*{1.}}
\put(32.0,0.0){\circle*{1.}}
\end{footnotesize}
\end{picture}}\hspace{0.5cm}
\subfigure[]{\begin{picture}(32.0,60.0)
\begin{footnotesize}
\linethickness{0.800000pt}
\put(3.2,60.0){\circle*{1.}}
\put(9.6,60.0){\circle*{1.}}
\put(16.0,60.0){\circle*{1.}}
\put(22.4,60.0){\circle*{1.}}
\put(28.800001,60.0){\circle*{1.}}
\put(0.0,54.0){\circle*{1.}}
\put(6.4,54.0){\circle*{1.}}
\put(12.8,54.0){\circle*{1.}}
\put(19.2,54.0){\circle*{1.}}
\put(25.6,54.0){\circle*{1.}}
\put(32.0,54.0){\circle*{1.}}
\put(3.2,48.0){\circle*{1.}}
\put(9.6,48.0){\circle*{1.}}
\put(16.0,48.0){\circle*{1.}}
\put(22.4,48.0){\circle*{1.}}
\put(28.800001,48.0){\circle*{1.}}
\put(0.0,42.0){\circle*{1.}}
\put(6.4,42.0){\circle*{1.}}
\put(12.8,42.0){\circle*{1.}}
\put(19.2,42.0){\circle*{1.}}
\put(25.6,42.0){\circle*{1.}}
\put(32.0,42.0){\circle*{1.}}
\put(3.2,36.0){\circle*{1.}}
\put(9.6,36.0){\circle*{1.}}
\put(16.0,36.0){\circle*{1.}}
\put(22.4,36.0){\circle*{1.}}
\put(28.800001,36.0){\circle*{1.}}
\put(0.0,30.0){\circle*{1.}}
\put(6.4,30.0){\circle*{1.}}
\put(12.8,30.0){\circle*{1.}}
\put(19.2,30.0){\circle*{1.}}
\put(25.6,30.0){\circle*{1.}}
\put(32.0,30.0){\circle*{1.}}
\put(3.2,24.0){\circle*{1.}}
\put(9.6,24.0){\circle{6.8}}
\put(16.0,24.0){\circle*{1.}}
\put(22.4,24.0){\circle*{1.}}
\put(28.800001,24.0){\circle*{1.}}
\put(0.0,18.0){\circle*{1.}}
\put(6.4,18.0){\circle*{1.}}
\put(12.8,18.0){\circle*{6.8}}
\put(19.2,18.0){\circle*{1.}}
\put(25.6,18.0){\circle*{1.}}
\put(32.0,18.0){\circle*{1.}}
\put(3.2,12.0){\circle*{1.}}
\put(9.6,12.0){\circle*{6.8}}
\put(16.0,12.0){\circle*{1.}}
\put(22.4,12.0){\circle*{1.}}
\put(28.800001,12.0){\circle*{1.}}
\put(0.0,6.0){\circle*{1.}}
\put(6.4,6.0){\circle*{1.}}
\put(12.8,6.0){\circle*{1.}}
\put(19.2,6.0){\circle*{1.}}
\put(25.6,6.0){\circle*{1.}}
\put(32.0,6.0){\circle*{1.}}
\put(3.2,0.0){\circle*{1.}}
\put(9.6,0.0){\circle*{1.}}
\put(16.0,0.0){\circle*{1.}}
\put(22.4,0.0){\circle*{1.}}
\put(28.800001,0.0){\circle*{1.}}
\end{footnotesize}
\end{picture} \hspace{0.3cm} \begin{picture}(32.0,60.0)
\begin{footnotesize}
\linethickness{0.800000pt}
\put(3.2,60.0){\circle*{1.}}
\put(9.6,60.0){\circle*{1.}}
\put(16.0,60.0){\circle*{1.}}
\put(22.4,60.0){\circle*{1.}}
\put(28.800001,60.0){\circle*{1.}}
\put(0.0,54.0){\circle*{1.}}
\put(6.4,54.0){\circle*{1.}}
\put(12.8,54.0){\circle*{1.}}
\put(19.2,54.0){\circle*{1.}}
\put(25.6,54.0){\circle*{1.}}
\put(32.0,54.0){\circle*{1.}}
\put(3.2,48.0){\circle*{1.}}
\put(9.6,48.0){\circle*{1.}}
\put(16.0,48.0){\circle*{1.}}
\put(22.4,48.0){\circle*{1.}}
\put(28.800001,48.0){\circle*{1.}}
\put(0.0,42.0){\circle*{1.}}
\put(6.4,42.0){\circle*{1.}}
\put(12.8,42.0){\circle*{1.}}
\put(19.2,42.0){\circle*{1.}}
\put(25.6,42.0){\circle*{1.}}
\put(32.0,42.0){\circle*{1.}}
\put(3.2,36.0){\circle*{1.}}
\put(9.6,36.0){\circle*{1.}}
\put(16.0,36.0){\circle*{1.}}
\put(22.4,36.0){\circle*{1.}}
\put(28.800001,36.0){\circle*{1.}}
\put(0.0,30.0){\circle*{1.}}
\put(6.4,30.0){\circle*{1.}}
\put(12.8,30.0){\circle*{1.}}
\put(19.2,30.0){\circle*{1.}}
\put(25.6,30.0){\circle*{1.}}
\put(32.0,30.0){\circle*{1.}}
\put(3.2,24.0){\circle*{1.}}
\put(9.6,24.0){\circle*{6.8}}
\put(16.0,24.0){\circle*{6.8}}
\put(22.4,24.0){\circle*{1.}}
\put(28.800001,24.0){\circle*{1.}}
\put(0.0,18.0){\circle*{6.8}}
\put(6.4,18.0){\circle*{1.}}
\put(12.8,18.0){\circle*{1.}}
\put(19.2,18.0){\circle{6.8}}
\put(25.6,18.0){\circle*{1.}}
\put(32.0,18.0){\circle*{1.}}
\put(3.2,12.0){\circle*{1.}}
\put(9.6,12.0){\circle*{1.}}
\put(16.0,12.0){\circle*{1.}}
\put(22.4,12.0){\circle*{1.}}
\put(28.800001,12.0){\circle*{1.}}
\put(0.0,6.0){\circle*{6.8}}
\put(6.4,6.0){\circle{6.8}}
\put(12.8,6.0){\circle{6.8}}
\put(19.2,6.0){\circle*{1.}}
\put(25.6,6.0){\circle*{1.}}
\put(32.0,6.0){\circle*{1.}}
\put(3.2,0.0){\circle*{1.}}
\put(9.6,0.0){\circle*{1.}}
\put(16.0,0.0){\circle*{1.}}
\put(22.4,0.0){\circle*{1.}}
\put(28.800001,0.0){\circle*{1.}}
\end{footnotesize}
\end{picture} \hspace{0.3cm} \begin{picture}(32.0,60.0)
\begin{footnotesize}
\linethickness{0.800000pt}
\put(3.2,60.0){\circle*{1.}}
\put(9.6,60.0){\circle*{1.}}
\put(16.0,60.0){\circle*{1.}}
\put(22.4,60.0){\circle*{1.}}
\put(28.800001,60.0){\circle*{1.}}
\put(0.0,54.0){\circle*{1.}}
\put(6.4,54.0){\circle*{1.}}
\put(12.8,54.0){\circle*{1.}}
\put(19.2,54.0){\circle*{1.}}
\put(25.6,54.0){\circle*{1.}}
\put(32.0,54.0){\circle*{1.}}
\put(3.2,48.0){\circle*{1.}}
\put(9.6,48.0){\circle*{1.}}
\put(16.0,48.0){\circle*{1.}}
\put(22.4,48.0){\circle*{1.}}
\put(28.800001,48.0){\circle*{1.}}
\put(0.0,42.0){\circle*{1.}}
\put(6.4,42.0){\circle*{1.}}
\put(12.8,42.0){\circle*{1.}}
\put(19.2,42.0){\circle*{1.}}
\put(25.6,42.0){\circle*{1.}}
\put(32.0,42.0){\circle*{1.}}
\put(3.2,36.0){\circle*{1.}}
\put(9.6,36.0){\circle*{1.}}
\put(16.0,36.0){\circle*{1.}}
\put(22.4,36.0){\circle*{1.}}
\put(28.800001,36.0){\circle*{1.}}
\put(0.0,30.0){\circle*{1.}}
\put(6.4,30.0){\circle{6.8}}
\put(12.8,30.0){\circle*{1.}}
\put(19.2,30.0){\circle{6.8}}
\put(25.6,30.0){\circle*{1.}}
\put(32.0,30.0){\circle*{1.}}
\put(3.2,24.0){\circle*{1.}}
\put(9.6,24.0){\circle*{1.}}
\put(16.0,24.0){\circle*{1.}}
\put(22.4,24.0){\circle*{6.8}}
\put(28.800001,24.0){\circle*{1.}}
\put(0.0,18.0){\circle*{1.}}
\put(6.4,18.0){\circle*{1.}}
\put(12.8,18.0){\circle*{1.}}
\put(19.2,18.0){\circle*{6.8}}
\put(25.6,18.0){\circle*{1.}}
\put(32.0,18.0){\circle*{1.}}
\put(3.2,12.0){\circle*{1.}}
\put(9.6,12.0){\circle*{1.}}
\put(16.0,12.0){\circle*{1.}}
\put(22.4,12.0){\circle*{1.}}
\put(28.800001,12.0){\circle*{1.}}
\put(0.0,6.0){\circle*{1.}}
\put(6.4,6.0){\circle*{6.8}}
\put(12.8,6.0){\circle*{1.}}
\put(19.2,6.0){\circle*{1.}}
\put(25.6,6.0){\circle*{1.}}
\put(32.0,6.0){\circle*{1.}}
\put(3.2,0.0){\circle*{6.8}}
\put(9.6,0.0){\circle*{1.}}
\put(16.0,0.0){\circle*{1.}}
\put(22.4,0.0){\circle*{1.}}
\put(28.800001,0.0){\circle*{1.}}
\end{footnotesize}
\end{picture} \hspace{0.3cm} \begin{picture}(32.0,60.0)
\begin{footnotesize}
\linethickness{0.800000pt}
\put(3.2,60.0){\circle*{1.}}
\put(9.6,60.0){\circle*{1.}}
\put(16.0,60.0){\circle*{1.}}
\put(22.4,60.0){\circle*{1.}}
\put(28.800001,60.0){\circle*{1.}}
\put(0.0,54.0){\circle*{1.}}
\put(6.4,54.0){\circle*{1.}}
\put(12.8,54.0){\circle*{1.}}
\put(19.2,54.0){\circle*{1.}}
\put(25.6,54.0){\circle*{1.}}
\put(32.0,54.0){\circle*{1.}}
\put(3.2,48.0){\circle*{1.}}
\put(9.6,48.0){\circle*{1.}}
\put(16.0,48.0){\circle*{1.}}
\put(22.4,48.0){\circle*{1.}}
\put(28.800001,48.0){\circle*{1.}}
\put(0.0,42.0){\circle*{1.}}
\put(6.4,42.0){\circle*{1.}}
\put(12.8,42.0){\circle*{1.}}
\put(19.2,42.0){\circle*{1.}}
\put(25.6,42.0){\circle*{1.}}
\put(32.0,42.0){\circle*{1.}}
\put(3.2,36.0){\circle*{1.}}
\put(9.6,36.0){\circle*{1.}}
\put(16.0,36.0){\circle*{1.}}
\put(22.4,36.0){\circle*{1.}}
\put(28.800001,36.0){\circle*{1.}}
\put(0.0,30.0){\circle*{1.}}
\put(6.4,30.0){\circle*{1.}}
\put(12.8,30.0){\circle*{1.}}
\put(19.2,30.0){\circle*{6.8}}
\put(25.6,30.0){\circle*{6.8}}
\put(32.0,30.0){\circle*{1.}}
\put(3.2,24.0){\circle*{1.}}
\put(9.6,24.0){\circle*{1.}}
\put(16.0,24.0){\circle*{1.}}
\put(22.4,24.0){\circle*{1.}}
\put(28.800001,24.0){\circle{6.8}}
\put(0.0,18.0){\circle*{1.}}
\put(6.4,18.0){\circle*{1.}}
\put(12.8,18.0){\circle{6.8}}
\put(19.2,18.0){\circle*{1.}}
\put(25.6,18.0){\circle*{1.}}
\put(32.0,18.0){\circle*{1.}}
\put(3.2,12.0){\circle{6.8}}
\put(9.6,12.0){\circle{6.8}}
\put(16.0,12.0){\circle{6.8}}
\put(22.4,12.0){\circle{6.8}}
\put(28.800001,12.0){\circle*{1.}}
\put(0.0,6.0){\circle*{1.}}
\put(6.4,6.0){\circle*{1.}}
\put(12.8,6.0){\circle{6.8}}
\put(19.2,6.0){\circle*{1.}}
\put(25.6,6.0){\circle*{1.}}
\put(32.0,6.0){\circle*{1.}}
\put(3.2,0.0){\circle*{1.}}
\put(9.6,0.0){\circle*{1.}}
\put(16.0,0.0){\circle*{1.}}
\put(22.4,0.0){\circle*{1.}}
\put(28.800001,0.0){\circle*{1.}}
\end{footnotesize}
\end{picture} \hspace{0.3cm} \begin{picture}(32.0,60.0)
\begin{footnotesize}
\linethickness{0.800000pt}
\put(3.2,60.0){\circle*{1.}}
\put(9.6,60.0){\circle*{1.}}
\put(16.0,60.0){\circle*{1.}}
\put(22.4,60.0){\circle*{1.}}
\put(28.800001,60.0){\circle*{1.}}
\put(0.0,54.0){\circle*{1.}}
\put(6.4,54.0){\circle*{1.}}
\put(12.8,54.0){\circle*{1.}}
\put(19.2,54.0){\circle*{1.}}
\put(25.6,54.0){\circle*{1.}}
\put(32.0,54.0){\circle*{1.}}
\put(3.2,48.0){\circle*{1.}}
\put(9.6,48.0){\circle*{1.}}
\put(16.0,48.0){\circle*{1.}}
\put(22.4,48.0){\circle*{1.}}
\put(28.800001,48.0){\circle*{1.}}
\put(0.0,42.0){\circle*{1.}}
\put(6.4,42.0){\circle*{1.}}
\put(12.8,42.0){\circle*{1.}}
\put(19.2,42.0){\circle*{1.}}
\put(25.6,42.0){\circle*{1.}}
\put(32.0,42.0){\circle*{1.}}
\put(3.2,36.0){\circle*{1.}}
\put(9.6,36.0){\circle*{1.}}
\put(16.0,36.0){\circle{6.8}}
\put(22.4,36.0){\circle*{1.}}
\put(28.800001,36.0){\circle{6.8}}
\put(0.0,30.0){\circle*{1.}}
\put(6.4,30.0){\circle*{1.}}
\put(12.8,30.0){\circle{6.8}}
\put(19.2,30.0){\circle*{1.}}
\put(25.6,30.0){\circle*{1.}}
\put(32.0,30.0){\circle*{6.8}}
\put(3.2,24.0){\circle*{1.}}
\put(9.6,24.0){\circle*{1.}}
\put(16.0,24.0){\circle*{6.8}}
\put(22.4,24.0){\circle*{1.}}
\put(28.800001,24.0){\circle*{6.8}}
\put(0.0,18.0){\circle*{1.}}
\put(6.4,18.0){\circle*{1.}}
\put(12.8,18.0){\circle*{1.}}
\put(19.2,18.0){\circle*{1.}}
\put(25.6,18.0){\circle*{1.}}
\put(32.0,18.0){\circle*{1.}}
\put(3.2,12.0){\circle*{1.}}
\put(9.6,12.0){\circle*{1.}}
\put(16.0,12.0){\circle*{1.}}
\put(22.4,12.0){\circle*{1.}}
\put(28.800001,12.0){\circle*{1.}}
\put(0.0,6.0){\circle*{6.8}}
\put(6.4,6.0){\circle*{1.}}
\put(12.8,6.0){\circle*{1.}}
\put(19.2,6.0){\circle*{1.}}
\put(25.6,6.0){\circle*{1.}}
\put(32.0,6.0){\circle*{1.}}
\put(3.2,0.0){\circle*{6.8}}
\put(9.6,0.0){\circle*{1.}}
\put(16.0,0.0){\circle*{1.}}
\put(22.4,0.0){\circle*{1.}}
\put(28.800001,0.0){\circle*{1.}}
\end{footnotesize}
\end{picture}}\hspace{0.5cm}
\subfigure[]{\begin{picture}(32.0,60.0)
\begin{footnotesize}
\linethickness{0.800000pt}
\put(3.2,60.0){\circle*{1.}}
\put(9.6,60.0){\circle*{1.}}
\put(16.0,60.0){\circle*{1.}}
\put(22.4,60.0){\circle*{1.}}
\put(28.800001,60.0){\circle*{1.}}
\put(0.0,54.0){\circle*{1.}}
\put(6.4,54.0){\circle*{1.}}
\put(12.8,54.0){\circle*{1.}}
\put(19.2,54.0){\circle*{1.}}
\put(25.6,54.0){\circle*{1.}}
\put(32.0,54.0){\circle*{1.}}
\put(3.2,48.0){\circle*{1.}}
\put(9.6,48.0){\circle*{1.}}
\put(16.0,48.0){\circle*{1.}}
\put(22.4,48.0){\circle*{1.}}
\put(28.800001,48.0){\circle*{1.}}
\put(0.0,42.0){\circle*{1.}}
\put(6.4,42.0){\circle*{1.}}
\put(12.8,42.0){\circle{6.8}}
\put(19.2,42.0){\circle*{1.}}
\put(25.6,42.0){\circle*{1.}}
\put(32.0,42.0){\circle*{1.}}
\put(3.2,36.0){\circle*{1.}}
\put(9.6,36.0){\circle*{1.}}
\put(16.0,36.0){\circle{6.8}}
\put(22.4,36.0){\circle*{1.}}
\put(28.800001,36.0){\circle*{1.}}
\put(0.0,30.0){\circle*{1.}}
\put(6.4,30.0){\circle*{1.}}
\put(12.8,30.0){\circle{6.8}}
\put(19.2,30.0){\circle*{1.}}
\put(25.6,30.0){\circle*{1.}}
\put(32.0,30.0){\circle*{1.}}
\put(3.2,24.0){\circle*{1.}}
\put(9.6,24.0){\circle*{1.}}
\put(16.0,24.0){\circle*{1.}}
\put(22.4,24.0){\circle*{1.}}
\put(28.800001,24.0){\circle*{1.}}
\put(0.0,18.0){\circle*{1.}}
\put(6.4,18.0){\circle*{1.}}
\put(12.8,18.0){\circle*{1.}}
\put(19.2,18.0){\circle*{1.}}
\put(25.6,18.0){\circle*{1.}}
\put(32.0,18.0){\circle*{1.}}
\put(3.2,12.0){\circle*{1.}}
\put(9.6,12.0){\circle*{1.}}
\put(16.0,12.0){\circle*{1.}}
\put(22.4,12.0){\circle*{1.}}
\put(28.800001,12.0){\circle*{1.}}
\put(0.0,6.0){\circle*{1.}}
\put(6.4,6.0){\circle*{1.}}
\put(12.8,6.0){\circle*{1.}}
\put(19.2,6.0){\circle*{1.}}
\put(25.6,6.0){\circle*{1.}}
\put(32.0,6.0){\circle*{1.}}
\put(3.2,0.0){\circle*{1.}}
\put(9.6,0.0){\circle*{1.}}
\put(16.0,0.0){\circle*{1.}}
\put(22.4,0.0){\circle*{1.}}
\put(28.800001,0.0){\circle*{1.}}
\end{footnotesize}
\end{picture} \hspace{0.3cm} \begin{picture}(32.0,60.0)
\begin{footnotesize}
\linethickness{0.800000pt}
\put(3.2,60.0){\circle*{1.}}
\put(9.6,60.0){\circle*{1.}}
\put(16.0,60.0){\circle*{1.}}
\put(22.4,60.0){\circle*{1.}}
\put(28.800001,60.0){\circle*{1.}}
\put(0.0,54.0){\circle*{1.}}
\put(6.4,54.0){\circle*{1.}}
\put(12.8,54.0){\circle*{1.}}
\put(19.2,54.0){\circle*{1.}}
\put(25.6,54.0){\circle*{1.}}
\put(32.0,54.0){\circle*{1.}}
\put(3.2,48.0){\circle*{1.}}
\put(9.6,48.0){\circle*{1.}}
\put(16.0,48.0){\circle*{1.}}
\put(22.4,48.0){\circle*{1.}}
\put(28.800001,48.0){\circle*{1.}}
\put(0.0,42.0){\circle*{1.}}
\put(6.4,42.0){\circle*{1.}}
\put(12.8,42.0){\circle*{1.}}
\put(19.2,42.0){\circle*{1.}}
\put(25.6,42.0){\circle*{1.}}
\put(32.0,42.0){\circle*{1.}}
\put(3.2,36.0){\circle*{1.}}
\put(9.6,36.0){\circle*{6.8}}
\put(16.0,36.0){\circle*{6.8}}
\put(22.4,36.0){\circle*{1.}}
\put(28.800001,36.0){\circle*{1.}}
\put(0.0,30.0){\circle*{1.}}
\put(6.4,30.0){\circle*{1.}}
\put(12.8,30.0){\circle*{1.}}
\put(19.2,30.0){\circle*{1.}}
\put(25.6,30.0){\circle*{1.}}
\put(32.0,30.0){\circle*{1.}}
\put(3.2,24.0){\circle*{1.}}
\put(9.6,24.0){\circle*{1.}}
\put(16.0,24.0){\circle*{1.}}
\put(22.4,24.0){\circle*{1.}}
\put(28.800001,24.0){\circle*{1.}}
\put(0.0,18.0){\circle*{1.}}
\put(6.4,18.0){\circle*{1.}}
\put(12.8,18.0){\circle*{1.}}
\put(19.2,18.0){\circle*{1.}}
\put(25.6,18.0){\circle*{1.}}
\put(32.0,18.0){\circle*{1.}}
\put(3.2,12.0){\circle*{1.}}
\put(9.6,12.0){\circle*{1.}}
\put(16.0,12.0){\circle*{1.}}
\put(22.4,12.0){\circle*{1.}}
\put(28.800001,12.0){\circle*{1.}}
\put(0.0,6.0){\circle*{1.}}
\put(6.4,6.0){\circle*{1.}}
\put(12.8,6.0){\circle*{1.}}
\put(19.2,6.0){\circle*{1.}}
\put(25.6,6.0){\circle*{1.}}
\put(32.0,6.0){\circle*{1.}}
\put(3.2,0.0){\circle*{1.}}
\put(9.6,0.0){\circle*{1.}}
\put(16.0,0.0){\circle*{1.}}
\put(22.4,0.0){\circle*{1.}}
\put(28.800001,0.0){\circle*{1.}}
\end{footnotesize}
\end{picture} \hspace{0.3cm} \begin{picture}(32.0,60.0)
\begin{footnotesize}
\linethickness{0.800000pt}
\put(3.2,60.0){\circle*{1.}}
\put(9.6,60.0){\circle*{1.}}
\put(16.0,60.0){\circle*{1.}}
\put(22.4,60.0){\circle*{1.}}
\put(28.800001,60.0){\circle*{1.}}
\put(0.0,54.0){\circle*{1.}}
\put(6.4,54.0){\circle*{1.}}
\put(12.8,54.0){\circle*{1.}}
\put(19.2,54.0){\circle*{1.}}
\put(25.6,54.0){\circle*{1.}}
\put(32.0,54.0){\circle*{1.}}
\put(3.2,48.0){\circle*{1.}}
\put(9.6,48.0){\circle*{1.}}
\put(16.0,48.0){\circle*{1.}}
\put(22.4,48.0){\circle*{1.}}
\put(28.800001,48.0){\circle*{1.}}
\put(0.0,42.0){\circle*{1.}}
\put(6.4,42.0){\circle{6.8}}
\put(12.8,42.0){\circle*{1.}}
\put(19.2,42.0){\circle{6.8}}
\put(25.6,42.0){\circle*{1.}}
\put(32.0,42.0){\circle*{1.}}
\put(3.2,36.0){\circle{6.8}}
\put(9.6,36.0){\circle*{1.}}
\put(16.0,36.0){\circle*{1.}}
\put(22.4,36.0){\circle{6.8}}
\put(28.800001,36.0){\circle*{1.}}
\put(0.0,30.0){\circle*{1.}}
\put(6.4,30.0){\circle{6.8}}
\put(12.8,30.0){\circle*{1.}}
\put(19.2,30.0){\circle{6.8}}
\put(25.6,30.0){\circle*{1.}}
\put(32.0,30.0){\circle*{1.}}
\put(3.2,24.0){\circle*{1.}}
\put(9.6,24.0){\circle*{1.}}
\put(16.0,24.0){\circle*{1.}}
\put(22.4,24.0){\circle*{1.}}
\put(28.800001,24.0){\circle*{1.}}
\put(0.0,18.0){\circle*{1.}}
\put(6.4,18.0){\circle*{1.}}
\put(12.8,18.0){\circle*{1.}}
\put(19.2,18.0){\circle*{1.}}
\put(25.6,18.0){\circle*{1.}}
\put(32.0,18.0){\circle*{1.}}
\put(3.2,12.0){\circle*{1.}}
\put(9.6,12.0){\circle*{1.}}
\put(16.0,12.0){\circle*{1.}}
\put(22.4,12.0){\circle*{1.}}
\put(28.800001,12.0){\circle*{1.}}
\put(0.0,6.0){\circle*{1.}}
\put(6.4,6.0){\circle*{1.}}
\put(12.8,6.0){\circle*{1.}}
\put(19.2,6.0){\circle*{1.}}
\put(25.6,6.0){\circle*{1.}}
\put(32.0,6.0){\circle*{1.}}
\put(3.2,0.0){\circle*{1.}}
\put(9.6,0.0){\circle*{1.}}
\put(16.0,0.0){\circle*{1.}}
\put(22.4,0.0){\circle*{1.}}
\put(28.800001,0.0){\circle*{1.}}
\end{footnotesize}
\end{picture} \hspace{0.3cm} \begin{picture}(32.0,60.0)
\begin{footnotesize}
\linethickness{0.800000pt}
\put(3.2,60.0){\circle*{1.}}
\put(9.6,60.0){\circle*{1.}}
\put(16.0,60.0){\circle*{1.}}
\put(22.4,60.0){\circle*{1.}}
\put(28.800001,60.0){\circle*{1.}}
\put(0.0,54.0){\circle*{1.}}
\put(6.4,54.0){\circle*{1.}}
\put(12.8,54.0){\circle*{1.}}
\put(19.2,54.0){\circle*{1.}}
\put(25.6,54.0){\circle*{1.}}
\put(32.0,54.0){\circle*{1.}}
\put(3.2,48.0){\circle*{1.}}
\put(9.6,48.0){\circle*{1.}}
\put(16.0,48.0){\circle*{1.}}
\put(22.4,48.0){\circle*{1.}}
\put(28.800001,48.0){\circle*{1.}}
\put(0.0,42.0){\circle*{1.}}
\put(6.4,42.0){\circle*{1.}}
\put(12.8,42.0){\circle*{1.}}
\put(19.2,42.0){\circle*{1.}}
\put(25.6,42.0){\circle*{1.}}
\put(32.0,42.0){\circle*{1.}}
\put(3.2,36.0){\circle*{1.}}
\put(9.6,36.0){\circle*{6.8}}
\put(16.0,36.0){\circle*{6.8}}
\put(22.4,36.0){\circle*{6.8}}
\put(28.800001,36.0){\circle*{1.}}
\put(0.0,30.0){\circle*{1.}}
\put(6.4,30.0){\circle*{1.}}
\put(12.8,30.0){\circle*{1.}}
\put(19.2,30.0){\circle*{1.}}
\put(25.6,30.0){\circle*{1.}}
\put(32.0,30.0){\circle*{1.}}
\put(3.2,24.0){\circle*{1.}}
\put(9.6,24.0){\circle*{1.}}
\put(16.0,24.0){\circle*{1.}}
\put(22.4,24.0){\circle*{1.}}
\put(28.800001,24.0){\circle*{1.}}
\put(0.0,18.0){\circle*{1.}}
\put(6.4,18.0){\circle*{1.}}
\put(12.8,18.0){\circle*{1.}}
\put(19.2,18.0){\circle*{1.}}
\put(25.6,18.0){\circle*{1.}}
\put(32.0,18.0){\circle*{1.}}
\put(3.2,12.0){\circle*{1.}}
\put(9.6,12.0){\circle*{1.}}
\put(16.0,12.0){\circle*{1.}}
\put(22.4,12.0){\circle*{1.}}
\put(28.800001,12.0){\circle*{1.}}
\put(0.0,6.0){\circle*{1.}}
\put(6.4,6.0){\circle*{1.}}
\put(12.8,6.0){\circle*{1.}}
\put(19.2,6.0){\circle*{1.}}
\put(25.6,6.0){\circle*{1.}}
\put(32.0,6.0){\circle*{1.}}
\put(3.2,0.0){\circle*{1.}}
\put(9.6,0.0){\circle*{1.}}
\put(16.0,0.0){\circle*{1.}}
\put(22.4,0.0){\circle*{1.}}
\put(28.800001,0.0){\circle*{1.}}
\end{footnotesize}
\end{picture} \hspace{0.3cm} \begin{picture}(32.0,60.0)
\begin{footnotesize}
\linethickness{0.800000pt}
\put(3.2,60.0){\circle*{1.}}
\put(9.6,60.0){\circle*{1.}}
\put(16.0,60.0){\circle*{1.}}
\put(22.4,60.0){\circle*{1.}}
\put(28.800001,60.0){\circle*{1.}}
\put(0.0,54.0){\circle*{1.}}
\put(6.4,54.0){\circle*{1.}}
\put(12.8,54.0){\circle*{1.}}
\put(19.2,54.0){\circle*{1.}}
\put(25.6,54.0){\circle*{1.}}
\put(32.0,54.0){\circle*{1.}}
\put(3.2,48.0){\circle*{1.}}
\put(9.6,48.0){\circle*{1.}}
\put(16.0,48.0){\circle*{1.}}
\put(22.4,48.0){\circle*{1.}}
\put(28.800001,48.0){\circle*{1.}}
\put(0.0,42.0){\circle*{1.}}
\put(6.4,42.0){\circle{6.8}}
\put(12.8,42.0){\circle*{1.}}
\put(19.2,42.0){\circle*{1.}}
\put(25.6,42.0){\circle{6.8}}
\put(32.0,42.0){\circle*{1.}}
\put(3.2,36.0){\circle{6.8}}
\put(9.6,36.0){\circle*{1.}}
\put(16.0,36.0){\circle*{1.}}
\put(22.4,36.0){\circle*{1.}}
\put(28.800001,36.0){\circle{6.8}}
\put(0.0,30.0){\circle*{1.}}
\put(6.4,30.0){\circle{6.8}}
\put(12.8,30.0){\circle*{1.}}
\put(19.2,30.0){\circle*{1.}}
\put(25.6,30.0){\circle{6.8}}
\put(32.0,30.0){\circle*{1.}}
\put(3.2,24.0){\circle*{1.}}
\put(9.6,24.0){\circle*{1.}}
\put(16.0,24.0){\circle*{1.}}
\put(22.4,24.0){\circle*{1.}}
\put(28.800001,24.0){\circle*{1.}}
\put(0.0,18.0){\circle*{1.}}
\put(6.4,18.0){\circle*{1.}}
\put(12.8,18.0){\circle*{1.}}
\put(19.2,18.0){\circle*{1.}}
\put(25.6,18.0){\circle*{1.}}
\put(32.0,18.0){\circle*{1.}}
\put(3.2,12.0){\circle*{1.}}
\put(9.6,12.0){\circle*{1.}}
\put(16.0,12.0){\circle*{1.}}
\put(22.4,12.0){\circle*{1.}}
\put(28.800001,12.0){\circle*{1.}}
\put(0.0,6.0){\circle*{1.}}
\put(6.4,6.0){\circle*{1.}}
\put(12.8,6.0){\circle*{1.}}
\put(19.2,6.0){\circle*{1.}}
\put(25.6,6.0){\circle*{1.}}
\put(32.0,6.0){\circle*{1.}}
\put(3.2,0.0){\circle*{1.}}
\put(9.6,0.0){\circle*{1.}}
\put(16.0,0.0){\circle*{1.}}
\put(22.4,0.0){\circle*{1.}}
\put(28.800001,0.0){\circle*{1.}}
\end{footnotesize}
\end{picture}}\hspace{0.5cm}
\caption{Examples of localizations observed in rules presented by glider-likehood matrices. State $A$ is shown by circle, state $B$ by solid disc.
(a)~glider propagating East; (b)~glider propagating North-East; (c)~puffer train propagating North-East; (d)~puffer train propagating East.}
\label{gliders}
\end{figure}

\begin{figure}
\centering
\includegraphics[width=0.5\linewidth]{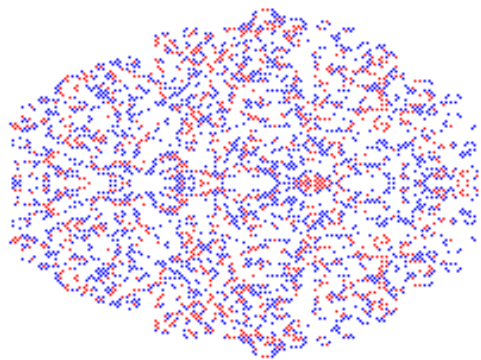}
\caption{An example of a typical configuration generated by glider-supporting rules. Gliders multiply during collisions, so the whole space is filled by visually disorganized but in reality highly ordered pattern. }
\label{disorderedpattern}
\end{figure}

Just a few examples of mobile localizations discovered are shown in Fig.~\ref{gliders}. Most typical localizations are gliders, which change their configuration every second time step (Fig.~\ref{gliders}ab), and puffer trains --- gliders leaving trail of breathers behind them (Fig.~\ref{gliders}cd). 
To complete this section we would like to stress that the evolutionary algorithm employed was tuned to select transition functions supporting localizations, not selecting rules with manageable dynamics of localizations. Therefore, those who will start experimenting with matrices presented in Fig.~\ref{likehoodmatrices} may be disappointed to see that, in most cases, development of an automaton from initial random configurations leads to 
disorderly looking configurations (even if the patch of initial stimulation was small enough). This is because gliders inhabit such spaces in abundance, they interact one with another, produce more gliders in result of their interaction, and populations of swarming gliders look like quasi-chaotic patterns for naked eyes (Fig.~\ref{disorderedpattern}).

\section{Quasi-chemical reaction}
\label{reactions}

\begin{figure}
\centering
\subfigure[$\mathbf{F}^S$]{\includegraphics[width=0.39\linewidth]{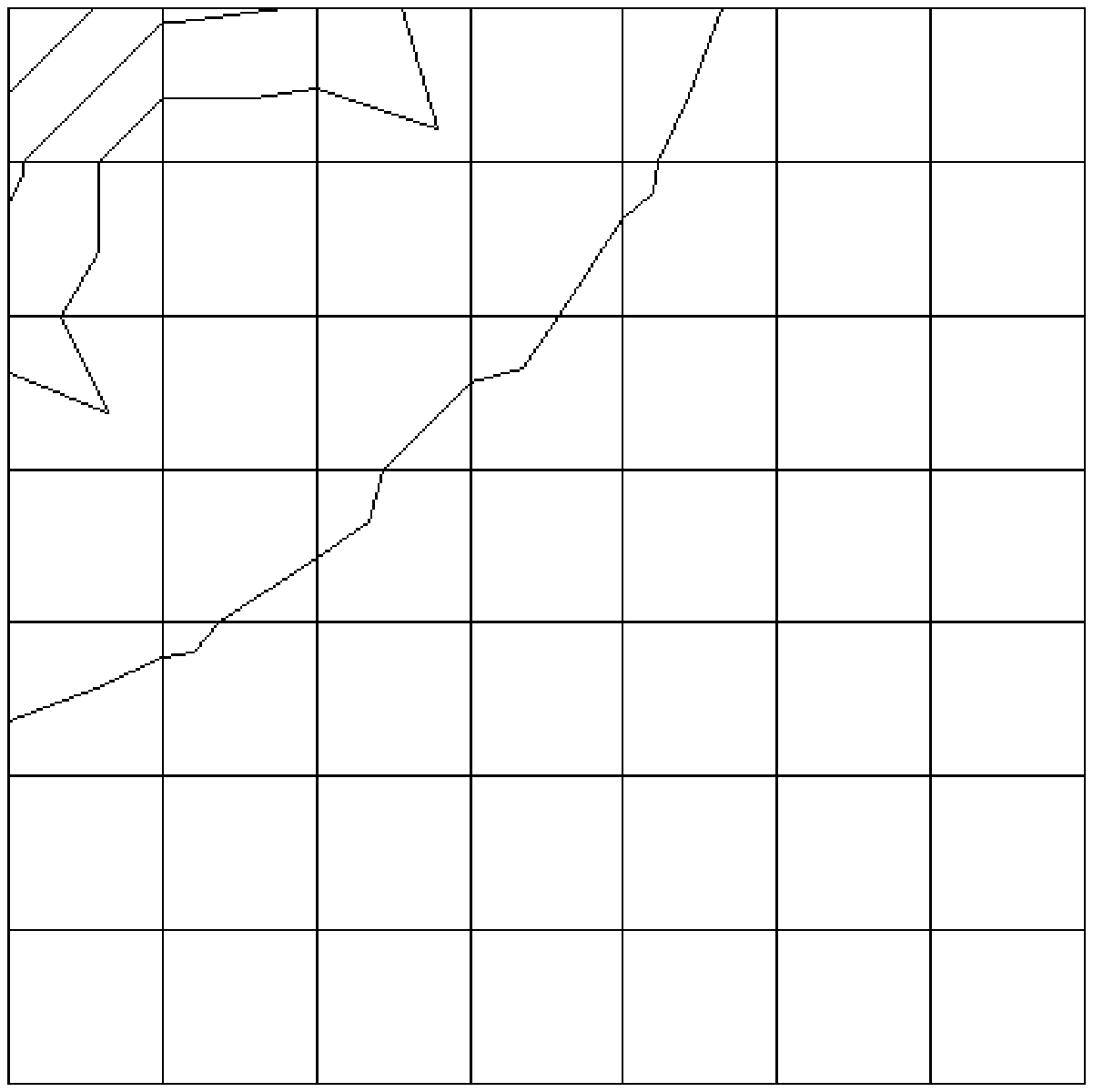}}
\subfigure[$\mathbf{F}^A$]{\includegraphics[width=0.39\linewidth]{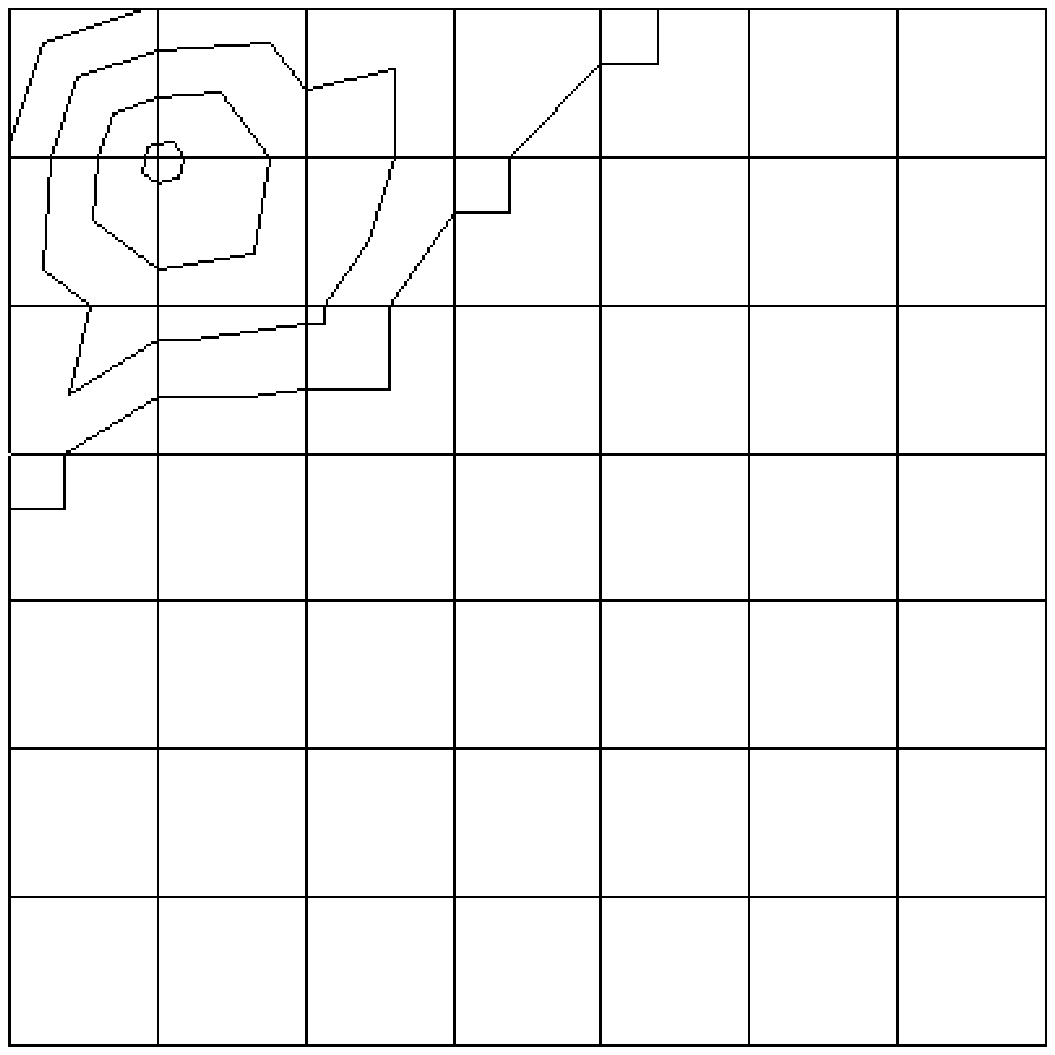}}
\subfigure[$\mathbf{F}^B$]{\includegraphics[width=0.39\linewidth]{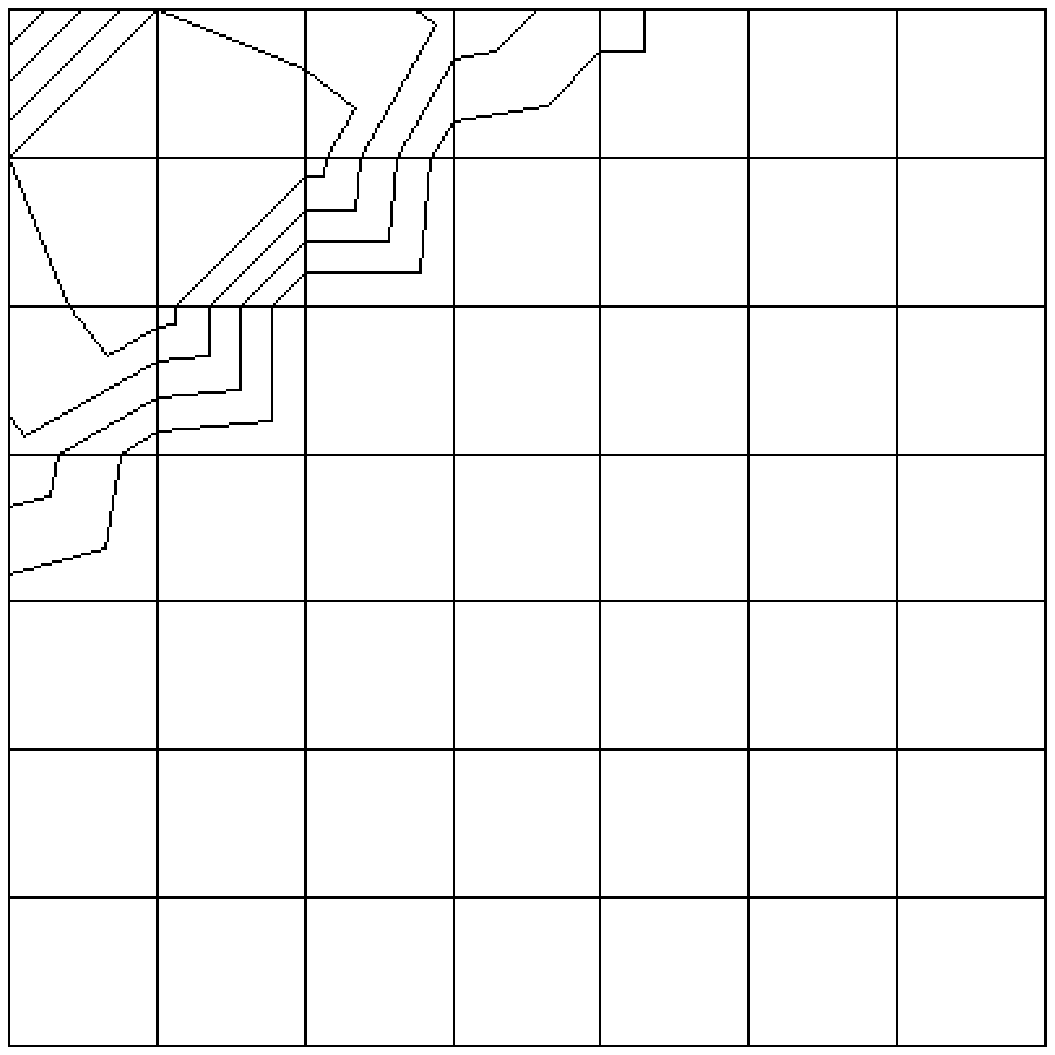}}
\subfigure[$\mathbf{F}^\#$]{\includegraphics[width=0.39\linewidth]{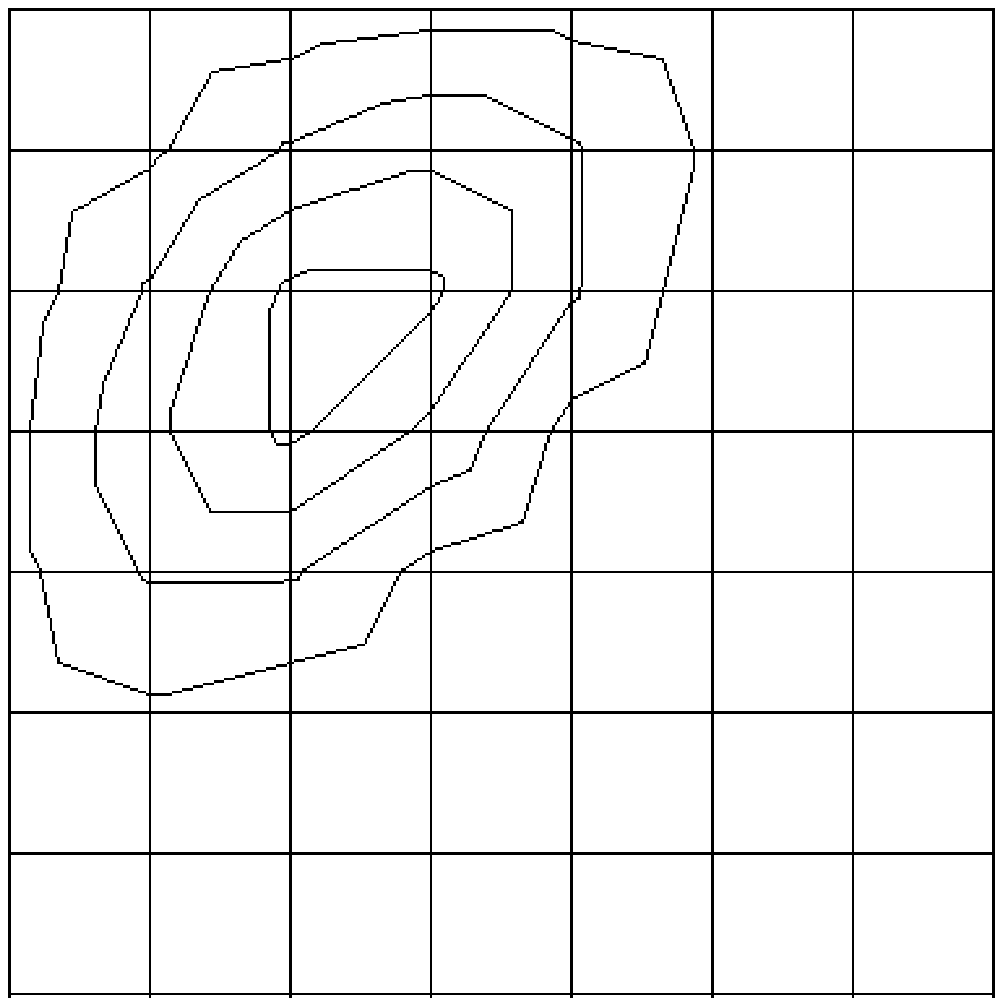}}
\caption{Isolines representation for glider likehood matrices. Number of states of reactant $A$ increases from top left corner to bottom left,
number of states of reactant $B$ increases from top left corner to top right one. In each case there is a single elevation. Approximate 
locations of elevations are $F_{00}^S$, $F_{11}^A$, $F_{11}^B$, and $F_{22}^\#$.}
\label{isolines}
\end{figure}


In order to characterize sets of possible quasi-chemical reactions,  which lead to the formation of 
stationary and mobile perturbations of reagent concentrations, we provide isoline representation of the glider likehood matrices in Fig.~\ref{isolines}. Based on the distributions shown in Fig.~\ref{isolines} we can speculate that

\begin{finding}
A probability, or rate, of dissociation of reactants decreases with the increase of number of the reactant molecules in each one local site of the medium. 
\end{finding}

See topology of isolines in Fig.~\ref{isolines}a. 

\begin{finding}
Reactions leading to production of reactants $A$ and $B$ are most likely to involve not more then four (usually between one and three) molecules of the reactants. 
\end{finding}

Distributions shown in Fig.~\ref{isolines}a and b reach their ground-zero levels (no reaction take place) for the sites where satisfying the condition: Sum of molecules $A$ and molecules $B$ does not exceed four.

\begin{finding}
Reactant $B$ is more likely to be produced during quasi-chemical reactions derived from $\mathbf{F}$-matrices.
\end{finding}

See Fig.~\ref{isolines}a and b to compare area of elevations.

Basing on these findings we generalize glider likehood matrices to a set of abstract, quasi-chemical, reactions as follows:

\begin{equation}
\begin{matrix}
   A + B 			\stackrel{1.}{\longrightarrow} 2 A 		&  \hspace{1.5cm} & A + 2 B 		\stackrel{0.4}{\longrightarrow} 2 A + B \cr
   2 A + B 		 \stackrel{0.1}{\longrightarrow} 3 A 	&   & 2 A + 2 B 	\stackrel{0.01}{\longrightarrow}	 3 A + B \cr
   3 A + B		 \stackrel{0.01}{\longrightarrow} 4 A &   & A + B 			 \stackrel{1.}{\longrightarrow} 2 B \cr
   A + 2 B 			\stackrel{0.1}{\longrightarrow} 3 B &   &  2 A + B 			\stackrel{0.05}{\longrightarrow} A + 2 B \cr
   B + S 				\stackrel{0.01.}{\longrightarrow} 2 B &   &  A 						\stackrel{0.054}{\longrightarrow} S  \cr
   B 						\stackrel{0.0015}{\longrightarrow} S  \cr   & & \cr
\end{matrix}
\end{equation}

\begin{figure}
\includegraphics[angle=-90,width=1.1\linewidth]{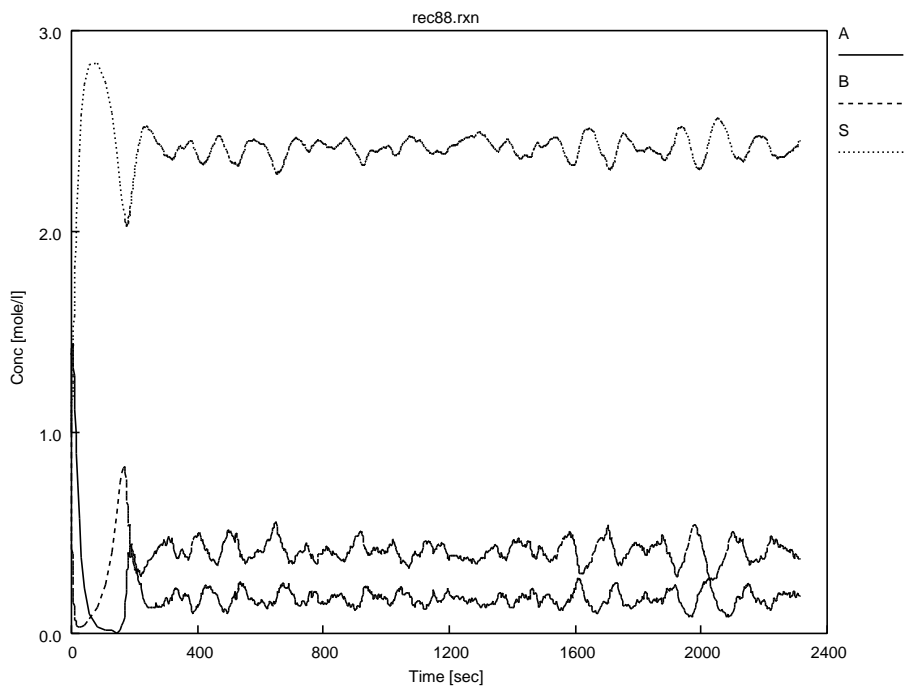}
\caption{Concentration dynamics of $A$, $B$ and $S$ species in a well-stirred reactor. 
Simulated on CKS (Chemical Kinetics Simulator, IBM) with the following parameters: pressure and temperature are kept constant, volume is not tracked. 100K particles were simulated. 
Initial concentration of each species is 1 mole/l.
}
\label{concentrations}
\end{figure}

A typical scenario of how the system (1) behaves in a well-stirred reactor is shown in Fig.~\ref{concentrations}. We have confirmed in the computational experiments that the reaction scheme developed represents an oscillatory chemical system, where concentration of substrate is significantly higher than concentrations of reactants $A$ and $B$. This indeed conforms with the nature of spreading localizations and pulsating behavior of generators of localizations, or glider guns. The presence of both types of reactants in a localization is essential for the majority of propagating localisations, which is identified by the similarity of the dynamic of concentration profiles of reactants $A$ and $B$ shown in  Fig.~\ref{concentrations}.

\section{Reductions of transitions functions}
\label{primitivisation}

What would be the behaviour of automata governed by the most likely cell-state transition rules?  In order to find an answer we decided to reduce matrices of 
glider likehoods to their most articulated forms. First, we have made matrices more symmetrical by the following procedure. If 
$|F'_{ij} - F''_{ji}|< \epsilon$ then we updated entries $(ij)$ and $(ji)$ as follows: 
$F'_{ij}=F''_{ji}=\left\lfloor \frac{1}{2}(F'_{ij}+F''_{ji}) \right\rfloor$. We do this because we assume that in evolutionary experiments reactants $A$ and $B$ have the same diffusion coefficients.

Then we perform the following operation. If $|F^A_{ij}-F^B_{ij}|>0.2$ and $|F^A_{ij}-F^S_{ij}|>0.2$, or $F^B_{ij}=0$ and $F^S_{ij}=0$ then $M^A_{ij}=1$. We also discard entries of $(0,5)$ and $(1,4)$ of the matrix $M^B$ as being negligibly small, and omit the matrix $F^\#$ out of 
consideration. After these operation, we got the following matrix

\begin{displaymath}
\mathbf{R}=
\begin{bmatrix}
0				&0					&0					&[0,1,2]	&0		&0		&0		&0 \cr		
0				&1					&[0,1,2]		&0				&0		&0		&0		&		\cr	
0				&[0,1,2] 	&[0,1]			&0				  &0		&0	  &			&			\cr	
[0,1,2]	&[0,1]			&0					&0				&0		&			&			&			\cr
[0,1]		&0					&0					&0				&			&			&			&	\cr
0				&0					&0					&					&			&			&	    &\cr
0				&0					&						&					&			&	    &     &\cr
0				&						&						&			    &     &     &     &\cr
\end{bmatrix}
\end{displaymath}

where $R_{ij}$ is a set of all possible states that a cell $x$ can take if its neighbourhood $u(x)$ has $i$ cells in state $A$ and $j$ cells in state $B$.  We found that 

\begin{finding}  Starting its development from a random configuration a 
cellular automaton governed by a local transition functions expressed by 
matrix $\mathbf{R}$ exhibits still and breathing localizations.  
\end{finding}

\begin{figure}
\centering
\subfigure[]{\begin{picture}(16.0,30.0)
\begin{footnotesize}
\linethickness{0.800000pt}
\put(3.2,30.0){\circle*{1.}}
\put(9.6,30.0){\circle*{1.}}
\put(16.0,30.0){\circle*{1.}}
\put(0.0,24.0){\circle*{1.}}
\put(6.4,24.0){\circle{6.8}}
\put(12.8,24.0){\circle*{1.}}
\put(3.2,18.0){\circle{6.8}}
\put(9.6,18.0){\circle{6.8}}
\put(16.0,18.0){\circle*{1.}}
\put(0.0,12.0){\circle*{1.}}
\put(6.4,12.0){\circle*{1.}}
\put(12.8,12.0){\circle*{1.}}
\put(3.2,6.0){\circle*{1.}}
\put(9.6,6.0){\circle*{1.}}
\put(16.0,6.0){\circle*{1.}}
\put(0.0,0.0){\circle*{1.}}
\put(6.4,0.0){\circle*{1.}}
\put(12.8,0.0){\circle*{1.}}
\end{footnotesize}
\end{picture} \hspace{0.3cm} \begin{picture}(16.0,30.0)
\begin{footnotesize}
\linethickness{0.800000pt}
\put(3.2,30.0){\circle*{1.}}
\put(9.6,30.0){\circle*{1.}}
\put(16.0,30.0){\circle*{1.}}
\put(0.0,24.0){\circle*{1.}}
\put(6.4,24.0){\circle*{6.8}}
\put(12.8,24.0){\circle*{1.}}
\put(3.2,18.0){\circle*{6.8}}
\put(9.6,18.0){\circle*{6.8}}
\put(16.0,18.0){\circle*{1.}}
\put(0.0,12.0){\circle*{1.}}
\put(6.4,12.0){\circle*{1.}}
\put(12.8,12.0){\circle*{1.}}
\put(3.2,6.0){\circle*{1.}}
\put(9.6,6.0){\circle*{1.}}
\put(16.0,6.0){\circle*{1.}}
\put(0.0,0.0){\circle*{1.}}
\put(6.4,0.0){\circle*{1.}}
\put(12.8,0.0){\circle*{1.}}
\end{footnotesize}
\end{picture} }\hspace{0.5cm}
\\
\subfigure[]{\begin{picture}(16.0,30.0)
\begin{footnotesize}
\linethickness{0.800000pt}
\put(0.0,30.0){\circle*{1.}}
\put(6.4,30.0){\circle*{1.}}
\put(12.8,30.0){\circle*{1.}}
\put(3.2,24.0){\circle*{1.}}
\put(9.6,24.0){\circle*{1.}}
\put(16.0,24.0){\circle*{1.}}
\put(0.0,18.0){\circle*{1.}}
\put(6.4,18.0){\circle{6.8}}
\put(12.8,18.0){\circle*{6.8}}
\put(3.2,12.0){\circle*{1.}}
\put(9.6,12.0){\circle*{6.8}}
\put(16.0,12.0){\circle{6.8}}
\put(0.0,6.0){\circle*{1.}}
\put(6.4,6.0){\circle*{1.}}
\put(12.8,6.0){\circle*{1.}}
\put(3.2,0.0){\circle*{1.}}
\put(9.6,0.0){\circle*{1.}}
\put(16.0,0.0){\circle*{1.}}
\end{footnotesize}
\end{picture}} \hspace{0.5cm}
\subfigure[]{\begin{picture}(16.0,30.0)
\begin{footnotesize}
\linethickness{0.800000pt}
\put(3.2,30.0){\circle*{1.}}
\put(9.6,30.0){\circle*{1.}}
\put(16.0,30.0){\circle*{1.}}
\put(0.0,24.0){\circle*{1.}}
\put(6.4,24.0){\circle*{6.8}}
\put(12.8,24.0){\circle*{6.8}}
\put(3.2,18.0){\circle*{1.}}
\put(9.6,18.0){\circle*{6.8}}
\put(16.0,18.0){\circle{6.8}}
\put(0.0,12.0){\circle*{1.}}
\put(6.4,12.0){\circle*{1.}}
\put(12.8,12.0){\circle*{1.}}
\put(3.2,6.0){\circle*{1.}}
\put(9.6,6.0){\circle*{1.}}
\put(16.0,6.0){\circle*{1.}}
\put(0.0,0.0){\circle*{1.}}
\put(6.4,0.0){\circle*{1.}}
\put(12.8,0.0){\circle*{1.}}
\end{footnotesize}
\end{picture}} \hspace{0.5cm}
\subfigure[]{\begin{picture}(16.0,30.0)
\begin{footnotesize}
\linethickness{0.800000pt}
\put(3.2,30.0){\circle*{1.}}
\put(9.6,30.0){\circle*{1.}}
\put(16.0,30.0){\circle*{1.}}
\put(0.0,24.0){\circle*{1.}}
\put(6.4,24.0){\circle{6.8}}
\put(12.8,24.0){\circle*{1.}}
\put(3.2,18.0){\circle*{6.8}}
\put(9.6,18.0){\circle*{6.8}}
\put(16.0,18.0){\circle*{1.}}
\put(0.0,12.0){\circle*{1.}}
\put(6.4,12.0){\circle{6.8}}
\put(12.8,12.0){\circle*{1.}}
\put(3.2,6.0){\circle*{1.}}
\put(9.6,6.0){\circle*{1.}}
\put(16.0,6.0){\circle*{1.}}
\put(0.0,0.0){\circle*{1.}}
\put(6.4,0.0){\circle*{1.}}
\put(12.8,0.0){\circle*{1.}}
\end{footnotesize}
\end{picture} \hspace{0.3cm} \begin{picture}(16.0,30.0)
\begin{footnotesize}
\linethickness{0.800000pt}
\put(3.2,30.0){\circle*{1.}}
\put(9.6,30.0){\circle*{1.}}
\put(16.0,30.0){\circle*{1.}}
\put(0.0,24.0){\circle*{1.}}
\put(6.4,24.0){\circle*{6.8}}
\put(12.8,24.0){\circle*{1.}}
\put(3.2,18.0){\circle*{6.8}}
\put(9.6,18.0){\circle*{6.8}}
\put(16.0,18.0){\circle*{1.}}
\put(0.0,12.0){\circle*{1.}}
\put(6.4,12.0){\circle*{6.8}}
\put(12.8,12.0){\circle*{1.}}
\put(3.2,6.0){\circle*{1.}}
\put(9.6,6.0){\circle*{1.}}
\put(16.0,6.0){\circle*{1.}}
\put(0.0,0.0){\circle*{1.}}
\put(6.4,0.0){\circle*{1.}}
\put(12.8,0.0){\circle*{1.}}
\end{footnotesize}
\end{picture} }\hspace{0.5cm}
\\
\subfigure[]{\begin{picture}(16.0,30.0)
\begin{footnotesize}
\linethickness{0.800000pt}
\put(0.0,30.0){\circle*{1.}}
\put(6.4,30.0){\circle*{1.}}
\put(12.8,30.0){\circle*{1.}}
\put(3.2,24.0){\circle*{1.}}
\put(9.6,24.0){\circle*{1.}}
\put(16.0,24.0){\circle*{1.}}
\put(0.0,18.0){\circle*{1.}}
\put(6.4,18.0){\circle{6.8}}
\put(12.8,18.0){\circle*{1.}}
\put(3.2,12.0){\circle*{1.}}
\put(9.6,12.0){\circle*{6.8}}
\put(16.0,12.0){\circle*{6.8}}
\put(0.0,6.0){\circle*{1.}}
\put(6.4,6.0){\circle{6.8}}
\put(12.8,6.0){\circle*{1.}}
\put(3.2,0.0){\circle*{1.}}
\put(9.6,0.0){\circle*{1.}}
\put(16.0,0.0){\circle*{1.}}
\end{footnotesize}
\end{picture} \hspace{0.3cm} \begin{picture}(16.0,30.0)
\begin{footnotesize}
\linethickness{0.800000pt}
\put(0.0,30.0){\circle*{1.}}
\put(6.4,30.0){\circle*{1.}}
\put(12.8,30.0){\circle*{1.}}
\put(3.2,24.0){\circle*{1.}}
\put(9.6,24.0){\circle*{1.}}
\put(16.0,24.0){\circle*{1.}}
\put(0.0,18.0){\circle*{1.}}
\put(6.4,18.0){\circle*{1.}}
\put(12.8,18.0){\circle{6.8}}
\put(3.2,12.0){\circle*{6.8}}
\put(9.6,12.0){\circle*{6.8}}
\put(16.0,12.0){\circle*{1.}}
\put(0.0,6.0){\circle*{1.}}
\put(6.4,6.0){\circle*{1.}}
\put(12.8,6.0){\circle{6.8}}
\put(3.2,0.0){\circle*{1.}}
\put(9.6,0.0){\circle*{1.}}
\put(16.0,0.0){\circle*{1.}}
\end{footnotesize}
\end{picture} }\hspace{0.5cm}
\subfigure[]{\begin{picture}(16.0,30.0)
\begin{footnotesize}
\linethickness{0.800000pt}
\put(0.0,30.0){\circle*{1.}}
\put(6.4,30.0){\circle*{1.}}
\put(12.8,30.0){\circle*{1.}}
\put(3.2,24.0){\circle*{6.8}}
\put(9.6,24.0){\circle*{1.}}
\put(16.0,24.0){\circle*{1.}}
\put(0.0,18.0){\circle*{1.}}
\put(6.4,18.0){\circle*{6.8}}
\put(12.8,18.0){\circle{6.8}}
\put(3.2,12.0){\circle*{6.8}}
\put(9.6,12.0){\circle*{1.}}
\put(16.0,12.0){\circle*{1.}}
\put(0.0,6.0){\circle*{1.}}
\put(6.4,6.0){\circle*{1.}}
\put(12.8,6.0){\circle*{1.}}
\put(3.2,0.0){\circle*{1.}}
\put(9.6,0.0){\circle*{1.}}
\put(16.0,0.0){\circle*{1.}}
\end{footnotesize}
\end{picture} \hspace{0.3cm} \begin{picture}(16.0,30.0)
\begin{footnotesize}
\linethickness{0.800000pt}
\put(0.0,30.0){\circle*{1.}}
\put(6.4,30.0){\circle*{1.}}
\put(12.8,30.0){\circle*{1.}}
\put(3.2,24.0){\circle*{1.}}
\put(9.6,24.0){\circle*{6.8}}
\put(16.0,24.0){\circle*{1.}}
\put(0.0,18.0){\circle{6.8}}
\put(6.4,18.0){\circle*{6.8}}
\put(12.8,18.0){\circle*{1.}}
\put(3.2,12.0){\circle*{1.}}
\put(9.6,12.0){\circle*{6.8}}
\put(16.0,12.0){\circle*{1.}}
\put(0.0,6.0){\circle*{1.}}
\put(6.4,6.0){\circle*{1.}}
\put(12.8,6.0){\circle*{1.}}
\put(3.2,0.0){\circle*{1.}}
\put(9.6,0.0){\circle*{1.}}
\put(16.0,0.0){\circle*{1.}}
\end{footnotesize}
\end{picture} }\hspace{0.5cm}
\subfigure[]{\begin{picture}(16.0,30.0)
\begin{footnotesize}
\linethickness{0.800000pt}
\put(3.2,30.0){\circle*{1.}}
\put(9.6,30.0){\circle*{1.}}
\put(16.0,30.0){\circle*{1.}}
\put(0.0,24.0){\circle*{1.}}
\put(6.4,24.0){\circle*{6.8}}
\put(12.8,24.0){\circle*{1.}}
\put(3.2,18.0){\circle*{1.}}
\put(9.6,18.0){\circle*{6.8}}
\put(16.0,18.0){\circle*{6.8}}
\put(0.0,12.0){\circle*{1.}}
\put(6.4,12.0){\circle*{6.8}}
\put(12.8,12.0){\circle*{1.}}
\put(3.2,6.0){\circle*{1.}}
\put(9.6,6.0){\circle*{1.}}
\put(16.0,6.0){\circle*{1.}}
\put(0.0,0.0){\circle*{1.}}
\put(6.4,0.0){\circle*{1.}}
\put(12.8,0.0){\circle*{1.}}
\end{footnotesize}
\end{picture} \hspace{0.3cm} \begin{picture}(16.0,30.0)
\begin{footnotesize}
\linethickness{0.800000pt}
\put(3.2,30.0){\circle*{1.}}
\put(9.6,30.0){\circle*{1.}}
\put(16.0,30.0){\circle*{1.}}
\put(0.0,24.0){\circle*{1.}}
\put(6.4,24.0){\circle*{1.}}
\put(12.8,24.0){\circle*{6.8}}
\put(3.2,18.0){\circle*{6.8}}
\put(9.6,18.0){\circle*{6.8}}
\put(16.0,18.0){\circle*{1.}}
\put(0.0,12.0){\circle*{1.}}
\put(6.4,12.0){\circle*{1.}}
\put(12.8,12.0){\circle*{6.8}}
\put(3.2,6.0){\circle*{1.}}
\put(9.6,6.0){\circle*{1.}}
\put(16.0,6.0){\circle*{1.}}
\put(0.0,0.0){\circle*{1.}}
\put(6.4,0.0){\circle*{1.}}
\put(12.8,0.0){\circle*{1.}}
\end{footnotesize}
\end{picture} }\hspace{0.5cm}
\subfigure[]{\begin{picture}(16.0,30.0)
\begin{footnotesize}
\linethickness{0.800000pt}
\put(3.2,30.0){\circle*{1.}}
\put(9.6,30.0){\circle*{1.}}
\put(16.0,30.0){\circle*{1.}}
\put(0.0,24.0){\circle*{1.}}
\put(6.4,24.0){\circle*{1.}}
\put(12.8,24.0){\circle{6.8}}
\put(3.2,18.0){\circle*{6.8}}
\put(9.6,18.0){\circle*{6.8}}
\put(16.0,18.0){\circle*{1.}}
\put(0.0,12.0){\circle*{1.}}
\put(6.4,12.0){\circle*{1.}}
\put(12.8,12.0){\circle*{6.8}}
\put(3.2,6.0){\circle*{1.}}
\put(9.6,6.0){\circle*{1.}}
\put(16.0,6.0){\circle*{1.}}
\put(0.0,0.0){\circle*{1.}}
\put(6.4,0.0){\circle*{1.}}
\put(12.8,0.0){\circle*{1.}}
\end{footnotesize}
\end{picture} \hspace{0.3cm} \begin{picture}(16.0,30.0)
\begin{footnotesize}
\linethickness{0.800000pt}
\put(3.2,30.0){\circle*{1.}}
\put(9.6,30.0){\circle*{1.}}
\put(16.0,30.0){\circle*{1.}}
\put(0.0,24.0){\circle*{1.}}
\put(6.4,24.0){\circle*{6.8}}
\put(12.8,24.0){\circle*{1.}}
\put(3.2,18.0){\circle*{1.}}
\put(9.6,18.0){\circle*{6.8}}
\put(16.0,18.0){\circle*{6.8}}
\put(0.0,12.0){\circle*{1.}}
\put(6.4,12.0){\circle{6.8}}
\put(12.8,12.0){\circle*{1.}}
\put(3.2,6.0){\circle*{1.}}
\put(9.6,6.0){\circle*{1.}}
\put(16.0,6.0){\circle*{1.}}
\put(0.0,0.0){\circle*{1.}}
\put(6.4,0.0){\circle*{1.}}
\put(12.8,0.0){\circle*{1.}}
\end{footnotesize}
\end{picture}\hspace{0.3cm}  }\hspace{0.5cm}
\\
\subfigure[]{\begin{picture}(16.0,30.0)
\begin{footnotesize}
\linethickness{0.800000pt}
\put(0.0,30.0){\circle*{1.}}
\put(6.4,30.0){\circle*{1.}}
\put(12.8,30.0){\circle*{1.}}
\put(3.2,24.0){\circle*{6.8}}
\put(9.6,24.0){\circle{6.8}}
\put(16.0,24.0){\circle*{1.}}
\put(0.0,18.0){\circle*{6.8}}
\put(6.4,18.0){\circle*{1.}}
\put(12.8,18.0){\circle{6.8}}
\put(3.2,12.0){\circle*{6.8}}
\put(9.6,12.0){\circle*{6.8}}
\put(16.0,12.0){\circle*{1.}}
\put(0.0,6.0){\circle*{1.}}
\put(6.4,6.0){\circle*{1.}}
\put(12.8,6.0){\circle*{1.}}
\put(3.2,0.0){\circle*{1.}}
\put(9.6,0.0){\circle*{1.}}
\put(16.0,0.0){\circle*{1.}}
\end{footnotesize}
\end{picture} \hspace{0.3cm} \begin{picture}(16.0,30.0)
\begin{footnotesize}
\linethickness{0.800000pt}
\put(0.0,30.0){\circle*{1.}}
\put(6.4,30.0){\circle*{1.}}
\put(12.8,30.0){\circle*{1.}}
\put(3.2,24.0){\circle*{6.8}}
\put(9.6,24.0){\circle*{6.8}}
\put(16.0,24.0){\circle*{1.}}
\put(0.0,18.0){\circle{6.8}}
\put(6.4,18.0){\circle*{1.}}
\put(12.8,18.0){\circle*{6.8}}
\put(3.2,12.0){\circle{6.8}}
\put(9.6,12.0){\circle*{6.8}}
\put(16.0,12.0){\circle*{1.}}
\put(0.0,6.0){\circle*{1.}}
\put(6.4,6.0){\circle*{1.}}
\put(12.8,6.0){\circle*{1.}}
\put(3.2,0.0){\circle*{1.}}
\put(9.6,0.0){\circle*{1.}}
\put(16.0,0.0){\circle*{1.}}
\end{footnotesize}
\end{picture} }\hspace{0.5cm}
\subfigure[]{\begin{picture}(32.0,60.0)
\begin{footnotesize}
\linethickness{0.800000pt}
\put(3.2,60.0){\circle*{1.}}
\put(9.6,60.0){\circle*{1.}}
\put(16.0,60.0){\circle*{1.}}
\put(22.4,60.0){\circle*{1.}}
\put(28.800001,60.0){\circle*{1.}}
\put(0.0,54.0){\circle*{1.}}
\put(6.4,54.0){\circle*{1.}}
\put(12.8,54.0){\circle*{1.}}
\put(19.2,54.0){\circle*{1.}}
\put(25.6,54.0){\circle*{1.}}
\put(32.0,54.0){\circle*{1.}}
\put(3.2,48.0){\circle*{1.}}
\put(9.6,48.0){\circle*{1.}}
\put(16.0,48.0){\circle*{1.}}
\put(22.4,48.0){\circle*{1.}}
\put(28.800001,48.0){\circle*{1.}}
\put(0.0,42.0){\circle*{1.}}
\put(6.4,42.0){\circle*{1.}}
\put(12.8,42.0){\circle*{1.}}
\put(19.2,42.0){\circle*{1.}}
\put(25.6,42.0){\circle*{1.}}
\put(32.0,42.0){\circle*{1.}}
\put(3.2,36.0){\circle*{1.}}
\put(9.6,36.0){\circle{6.8}}
\put(16.0,36.0){\circle{6.8}}
\put(22.4,36.0){\circle*{6.8}}
\put(28.800001,36.0){\circle*{1.}}
\put(0.0,30.0){\circle*{1.}}
\put(6.4,30.0){\circle*{6.8}}
\put(12.8,30.0){\circle*{1.}}
\put(19.2,30.0){\circle*{1.}}
\put(25.6,30.0){\circle*{6.8}}
\put(32.0,30.0){\circle*{1.}}
\put(3.2,24.0){\circle*{1.}}
\put(9.6,24.0){\circle{6.8}}
\put(16.0,24.0){\circle{6.8}}
\put(22.4,24.0){\circle*{6.8}}
\put(28.800001,24.0){\circle*{1.}}
\put(0.0,18.0){\circle*{1.}}
\put(6.4,18.0){\circle*{1.}}
\put(12.8,18.0){\circle*{1.}}
\put(19.2,18.0){\circle*{1.}}
\put(25.6,18.0){\circle*{1.}}
\put(32.0,18.0){\circle*{1.}}
\put(3.2,12.0){\circle*{1.}}
\put(9.6,12.0){\circle*{1.}}
\put(16.0,12.0){\circle*{1.}}
\put(22.4,12.0){\circle*{1.}}
\put(28.800001,12.0){\circle*{1.}}
\put(0.0,6.0){\circle*{1.}}
\put(6.4,6.0){\circle*{1.}}
\put(12.8,6.0){\circle*{1.}}
\put(19.2,6.0){\circle*{1.}}
\put(25.6,6.0){\circle*{1.}}
\put(32.0,6.0){\circle*{1.}}
\put(3.2,0.0){\circle*{1.}}
\put(9.6,0.0){\circle*{1.}}
\put(16.0,0.0){\circle*{1.}}
\put(22.4,0.0){\circle*{1.}}
\put(28.800001,0.0){\circle*{1.}}
\end{footnotesize}
\end{picture} \hspace{0.3cm} \begin{picture}(32.0,60.0)
\begin{footnotesize}
\linethickness{0.800000pt}
\put(3.2,60.0){\circle*{1.}}
\put(9.6,60.0){\circle*{1.}}
\put(16.0,60.0){\circle*{1.}}
\put(22.4,60.0){\circle*{1.}}
\put(28.800001,60.0){\circle*{1.}}
\put(0.0,54.0){\circle*{1.}}
\put(6.4,54.0){\circle*{1.}}
\put(12.8,54.0){\circle*{1.}}
\put(19.2,54.0){\circle*{1.}}
\put(25.6,54.0){\circle*{1.}}
\put(32.0,54.0){\circle*{1.}}
\put(3.2,48.0){\circle*{1.}}
\put(9.6,48.0){\circle*{1.}}
\put(16.0,48.0){\circle*{1.}}
\put(22.4,48.0){\circle*{1.}}
\put(28.800001,48.0){\circle*{1.}}
\put(0.0,42.0){\circle*{1.}}
\put(6.4,42.0){\circle*{1.}}
\put(12.8,42.0){\circle*{1.}}
\put(19.2,42.0){\circle*{1.}}
\put(25.6,42.0){\circle*{1.}}
\put(32.0,42.0){\circle*{1.}}
\put(3.2,36.0){\circle*{1.}}
\put(9.6,36.0){\circle*{6.8}}
\put(16.0,36.0){\circle*{6.8}}
\put(22.4,36.0){\circle{6.8}}
\put(28.800001,36.0){\circle*{1.}}
\put(0.0,30.0){\circle*{1.}}
\put(6.4,30.0){\circle*{6.8}}
\put(12.8,30.0){\circle*{1.}}
\put(19.2,30.0){\circle*{1.}}
\put(25.6,30.0){\circle*{6.8}}
\put(32.0,30.0){\circle*{1.}}
\put(3.2,24.0){\circle*{1.}}
\put(9.6,24.0){\circle*{6.8}}
\put(16.0,24.0){\circle*{6.8}}
\put(22.4,24.0){\circle{6.8}}
\put(28.800001,24.0){\circle*{1.}}
\put(0.0,18.0){\circle*{1.}}
\put(6.4,18.0){\circle*{1.}}
\put(12.8,18.0){\circle*{1.}}
\put(19.2,18.0){\circle*{1.}}
\put(25.6,18.0){\circle*{1.}}
\put(32.0,18.0){\circle*{1.}}
\put(3.2,12.0){\circle*{1.}}
\put(9.6,12.0){\circle*{1.}}
\put(16.0,12.0){\circle*{1.}}
\put(22.4,12.0){\circle*{1.}}
\put(28.800001,12.0){\circle*{1.}}
\put(0.0,6.0){\circle*{1.}}
\put(6.4,6.0){\circle*{1.}}
\put(12.8,6.0){\circle*{1.}}
\put(19.2,6.0){\circle*{1.}}
\put(25.6,6.0){\circle*{1.}}
\put(32.0,6.0){\circle*{1.}}
\put(3.2,0.0){\circle*{1.}}
\put(9.6,0.0){\circle*{1.}}
\put(16.0,0.0){\circle*{1.}}
\put(22.4,0.0){\circle*{1.}}
\put(28.800001,0.0){\circle*{1.}}
\end{footnotesize}
\end{picture}
 \hspace{0.3cm} \begin{picture}(32.0,60.0)
\begin{footnotesize}
\linethickness{0.800000pt}
\put(3.2,60.0){\circle*{1.}}
\put(9.6,60.0){\circle*{1.}}
\put(16.0,60.0){\circle*{1.}}
\put(22.4,60.0){\circle*{1.}}
\put(28.800001,60.0){\circle*{1.}}
\put(0.0,54.0){\circle*{1.}}
\put(6.4,54.0){\circle*{1.}}
\put(12.8,54.0){\circle*{1.}}
\put(19.2,54.0){\circle*{1.}}
\put(25.6,54.0){\circle*{1.}}
\put(32.0,54.0){\circle*{1.}}
\put(3.2,48.0){\circle*{1.}}
\put(9.6,48.0){\circle*{1.}}
\put(16.0,48.0){\circle*{1.}}
\put(22.4,48.0){\circle*{1.}}
\put(28.800001,48.0){\circle*{1.}}
\put(0.0,42.0){\circle*{1.}}
\put(6.4,42.0){\circle*{1.}}
\put(12.8,42.0){\circle*{1.}}
\put(19.2,42.0){\circle*{1.}}
\put(25.6,42.0){\circle*{1.}}
\put(32.0,42.0){\circle*{1.}}
\put(3.2,36.0){\circle*{1.}}
\put(9.6,36.0){\circle*{6.8}}
\put(16.0,36.0){\circle{6.8}}
\put(22.4,36.0){\circle{6.8}}
\put(28.800001,36.0){\circle*{1.}}
\put(0.0,30.0){\circle*{1.}}
\put(6.4,30.0){\circle*{6.8}}
\put(12.8,30.0){\circle*{1.}}
\put(19.2,30.0){\circle*{1.}}
\put(25.6,30.0){\circle*{6.8}}
\put(32.0,30.0){\circle*{1.}}
\put(3.2,24.0){\circle*{1.}}
\put(9.6,24.0){\circle*{6.8}}
\put(16.0,24.0){\circle{6.8}}
\put(22.4,24.0){\circle{6.8}}
\put(28.800001,24.0){\circle*{1.}}
\put(0.0,18.0){\circle*{1.}}
\put(6.4,18.0){\circle*{1.}}
\put(12.8,18.0){\circle*{1.}}
\put(19.2,18.0){\circle*{1.}}
\put(25.6,18.0){\circle*{1.}}
\put(32.0,18.0){\circle*{1.}}
\put(3.2,12.0){\circle*{1.}}
\put(9.6,12.0){\circle*{1.}}
\put(16.0,12.0){\circle*{1.}}
\put(22.4,12.0){\circle*{1.}}
\put(28.800001,12.0){\circle*{1.}}
\put(0.0,6.0){\circle*{1.}}
\put(6.4,6.0){\circle*{1.}}
\put(12.8,6.0){\circle*{1.}}
\put(19.2,6.0){\circle*{1.}}
\put(25.6,6.0){\circle*{1.}}
\put(32.0,6.0){\circle*{1.}}
\put(3.2,0.0){\circle*{1.}}
\put(9.6,0.0){\circle*{1.}}
\put(16.0,0.0){\circle*{1.}}
\put(22.4,0.0){\circle*{1.}}
\put(28.800001,0.0){\circle*{1.}}
\end{footnotesize}
\end{picture} }\hspace{0.5cm}
\subfigure[]{\begin{picture}(32.0,60.0)
\begin{footnotesize}
\linethickness{0.800000pt}
\put(0.0,60.0){\circle*{1.}}
\put(6.4,60.0){\circle*{1.}}
\put(12.8,60.0){\circle*{1.}}
\put(19.2,60.0){\circle*{1.}}
\put(25.6,60.0){\circle*{1.}}
\put(32.0,60.0){\circle*{1.}}
\put(3.2,54.0){\circle*{1.}}
\put(9.6,54.0){\circle*{1.}}
\put(16.0,54.0){\circle*{1.}}
\put(22.4,54.0){\circle*{1.}}
\put(28.800001,54.0){\circle*{1.}}
\put(0.0,48.0){\circle*{1.}}
\put(6.4,48.0){\circle*{1.}}
\put(12.8,48.0){\circle*{1.}}
\put(19.2,48.0){\circle*{1.}}
\put(25.6,48.0){\circle*{1.}}
\put(32.0,48.0){\circle*{1.}}
\put(3.2,42.0){\circle*{1.}}
\put(9.6,42.0){\circle*{1.}}
\put(16.0,42.0){\circle*{6.8}}
\put(22.4,42.0){\circle*{6.8}}
\put(28.800001,42.0){\circle*{1.}}
\put(0.0,36.0){\circle*{1.}}
\put(6.4,36.0){\circle*{1.}}
\put(12.8,36.0){\circle*{6.8}}
\put(19.2,36.0){\circle*{1.}}
\put(25.6,36.0){\circle*{6.8}}
\put(32.0,36.0){\circle*{1.}}
\put(3.2,30.0){\circle*{1.}}
\put(9.6,30.0){\circle*{6.8}}
\put(16.0,30.0){\circle*{1.}}
\put(22.4,30.0){\circle*{6.8}}
\put(28.800001,30.0){\circle*{1.}}
\put(0.0,24.0){\circle*{1.}}
\put(6.4,24.0){\circle*{6.8}}
\put(12.8,24.0){\circle*{1.}}
\put(19.2,24.0){\circle*{6.8}}
\put(25.6,24.0){\circle*{1.}}
\put(32.0,24.0){\circle*{1.}}
\put(3.2,18.0){\circle*{1.}}
\put(9.6,18.0){\circle*{6.8}}
\put(16.0,18.0){\circle*{6.8}}
\put(22.4,18.0){\circle*{1.}}
\put(28.800001,18.0){\circle*{1.}}
\put(0.0,12.0){\circle*{1.}}
\put(6.4,12.0){\circle*{1.}}
\put(12.8,12.0){\circle*{1.}}
\put(19.2,12.0){\circle*{1.}}
\put(25.6,12.0){\circle*{1.}}
\put(32.0,12.0){\circle*{1.}}
\put(3.2,6.0){\circle*{1.}}
\put(9.6,6.0){\circle*{1.}}
\put(16.0,6.0){\circle*{1.}}
\put(22.4,6.0){\circle*{1.}}
\put(28.800001,6.0){\circle*{1.}}
\put(0.0,0.0){\circle*{1.}}
\put(6.4,0.0){\circle*{1.}}
\put(12.8,0.0){\circle*{1.}}
\put(19.2,0.0){\circle*{1.}}
\put(25.6,0.0){\circle*{1.}}
\put(32.0,0.0){\circle*{1.}}
\end{footnotesize}
\end{picture} \hspace{0.3cm} \begin{picture}(32.0,60.0)
\begin{footnotesize}
\linethickness{0.800000pt}
\put(0.0,60.0){\circle*{1.}}
\put(6.4,60.0){\circle*{1.}}
\put(12.8,60.0){\circle*{1.}}
\put(19.2,60.0){\circle*{1.}}
\put(25.6,60.0){\circle*{1.}}
\put(32.0,60.0){\circle*{1.}}
\put(3.2,54.0){\circle*{1.}}
\put(9.6,54.0){\circle*{1.}}
\put(16.0,54.0){\circle*{1.}}
\put(22.4,54.0){\circle*{1.}}
\put(28.800001,54.0){\circle*{1.}}
\put(0.0,48.0){\circle*{1.}}
\put(6.4,48.0){\circle*{1.}}
\put(12.8,48.0){\circle*{1.}}
\put(19.2,48.0){\circle*{1.}}
\put(25.6,48.0){\circle*{1.}}
\put(32.0,48.0){\circle*{1.}}
\put(3.2,42.0){\circle*{1.}}
\put(9.6,42.0){\circle*{1.}}
\put(16.0,42.0){\circle*{6.8}}
\put(22.4,42.0){\circle*{6.8}}
\put(28.800001,42.0){\circle*{1.}}
\put(0.0,36.0){\circle*{1.}}
\put(6.4,36.0){\circle*{1.}}
\put(12.8,36.0){\circle*{6.8}}
\put(19.2,36.0){\circle*{1.}}
\put(25.6,36.0){\circle*{6.8}}
\put(32.0,36.0){\circle*{1.}}
\put(3.2,30.0){\circle*{1.}}
\put(9.6,30.0){\circle*{6.8}}
\put(16.0,30.0){\circle*{6.8}}
\put(22.4,30.0){\circle*{6.8}}
\put(28.800001,30.0){\circle*{1.}}
\put(0.0,24.0){\circle*{1.}}
\put(6.4,24.0){\circle*{6.8}}
\put(12.8,24.0){\circle*{1.}}
\put(19.2,24.0){\circle*{6.8}}
\put(25.6,24.0){\circle*{1.}}
\put(32.0,24.0){\circle*{1.}}
\put(3.2,18.0){\circle*{1.}}
\put(9.6,18.0){\circle*{6.8}}
\put(16.0,18.0){\circle*{6.8}}
\put(22.4,18.0){\circle*{1.}}
\put(28.800001,18.0){\circle*{1.}}
\put(0.0,12.0){\circle*{1.}}
\put(6.4,12.0){\circle*{1.}}
\put(12.8,12.0){\circle*{1.}}
\put(19.2,12.0){\circle*{1.}}
\put(25.6,12.0){\circle*{1.}}
\put(32.0,12.0){\circle*{1.}}
\put(3.2,6.0){\circle*{1.}}
\put(9.6,6.0){\circle*{1.}}
\put(16.0,6.0){\circle*{1.}}
\put(22.4,6.0){\circle*{1.}}
\put(28.800001,6.0){\circle*{1.}}
\put(0.0,0.0){\circle*{1.}}
\put(6.4,0.0){\circle*{1.}}
\put(12.8,0.0){\circle*{1.}}
\put(19.2,0.0){\circle*{1.}}
\put(25.6,0.0){\circle*{1.}}
\put(32.0,0.0){\circle*{1.}}
\end{footnotesize}
\end{picture}
}\hspace{0.5cm}
\caption{Stationary localizations typically observed in automata
governed by cell-state transition rules from class $\mathbf{R}$. State $A$ is shown by a circle, state $B$ by a solid disc.}
\label{flocalizations}
\end{figure}
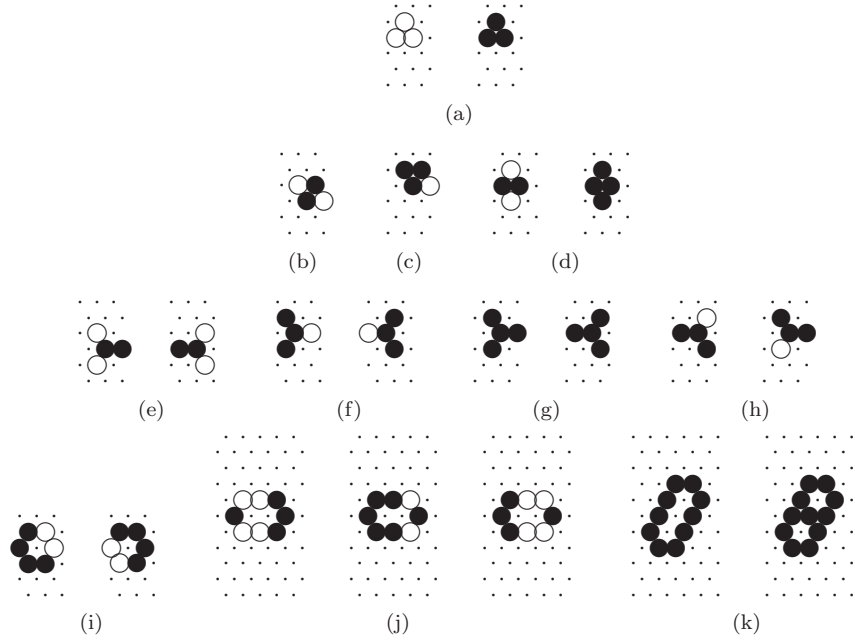

Most common localizations observed in cellular automata with local transition rules from the class $\mathbf{R}$ are shown in 
Fig.~\ref{flocalizations}. Minimal weight localizations are still triples of cells in state $A$ or $B$, or triples switching between $A$ and $B$~(Fig.~\ref{flocalizations}a). These are followed by compact still patterns of four cells in $A$ or $B$ states, see examples in~Fig.~\ref{flocalizations}bc, switching compact patterns of four non-resting states~Fig.~\ref{flocalizations}d;  and, wish-bone shaped clusters of $A$ and $B$ states Fig.~\ref{flocalizations}e--h. 
The next heavier common localizations are the switching patterns 
of six~Fig.~\ref{flocalizations}i and eighth~Fig.~\ref{flocalizations}j  non-resting states, and, the breathing localization~Fig.~\ref{flocalizations}k oscillating between 12 and 13 non-resting states.

\section{Conclusion}
\label{discussions}

We have hybridized the paradigm of reaction-diffusion cellular 
automata~\cite{adamatzky_hexgliders_2006,wuensche_adamatzky_2006,adamatzky_wuensche_2007} with evolutionary techniques of 
breeding glider-supporting rules~\cite{sapin_2003,sapin_2003a,sapin_2004,sapin_2007} to statistically evaluate the set of all 
possible totalistic cell-state transition functions which support mobile and stationary localizations. We calculated exact
structures of glider likehood matrices and interpreted them in terms of abstract reactions. We demonstrated that
quasi-chemical systems derived from glider-supporting rules exhibit classical dynamics of excitable chemical systems. We obtained 
computational experiment evidences that by reducing glider likehood matrices to their strong components we obtain a set of 
local transition rules that exhibit exclusively stationary localizations. Results of the research undertaken provide a 
priceless tool for designing collision-based computing schemes in spatially extended non-linear systems.

\end{document}